\documentclass[10pt,twocolumn,letterpaper]{article}

\usepackage[accsupp]{axessibility}
\usepackage[pagenumbers]{cvpr} 

\definecolor{cvprblue}{rgb}{0.21,0.49,0.74}
\usepackage[pagebackref,breaklinks,colorlinks,allcolors=cvprblue]{hyperref}

\def\paperID{3633} 
\def\confName{CVPR}
\def\confYear{2025}
\usepackage{multirow}
\usepackage{natbib}
\usepackage{pifont}
\usepackage{array}
\title{Adaptive Dropout: Unleashing Dropout across Layers for Generalizable Image Super-Resolution}

\author{Hang Xu \quad  Wei Yu \quad Jiangtong Tan \quad Zhen Zou \quad Feng Zhao\thanks{Corresponding author}\\
MoE Key Laboratory of Brain-inspired Intelligent Perception and Cognition\\
University of Science and Technology of China\\
{\tt\small \{xuhang0609 patrick914y, jttan, zouzhen\}@mail.ustc.edu.cn, fzhao956@ustc.edu.cn}
}

\begin{document}
\maketitle
\begin{abstract}
Blind Super-Resolution (blind SR) aims to enhance the model's generalization ability with unknown degradation, yet it still encounters severe overfitting issues. Some previous methods inspired by dropout, which enhances generalization by regularizing features, have shown promising results in blind SR. Nevertheless, these methods focus solely on regularizing features before the final layer and overlook the need for generalization in features at intermediate layers. Without explicit regularization of features at intermediate layers, the blind SR network struggles to obtain well-generalized feature representations. However, the key challenge is that directly applying dropout to intermediate layers leads to a significant performance drop, which we attribute to the inconsistency in training-testing and across layers it introduced. Therefore, we propose Adaptive Dropout, a new regularization method for blind SR models, which mitigates the inconsistency and facilitates application across intermediate layers of networks. Specifically, for training-testing inconsistency, we re-design the form of dropout and integrate the features before and after dropout adaptively. For inconsistency in generalization requirements across different layers, we innovatively design an adaptive training strategy to strengthen feature propagation by layer-wise annealing. Experimental results show that our method outperforms all past regularization methods on both synthetic and real-world benchmark datasets, also highly effective in other image restoration tasks. Code is available at \href{https://github.com/xuhang07/Adpative-Dropout}{https://github.com/xuhang07/Adpative-Dropout}.
\end{abstract}
\vspace{-4mm}    
\begin{figure}[t]
\centering
\includegraphics[width=1\linewidth]{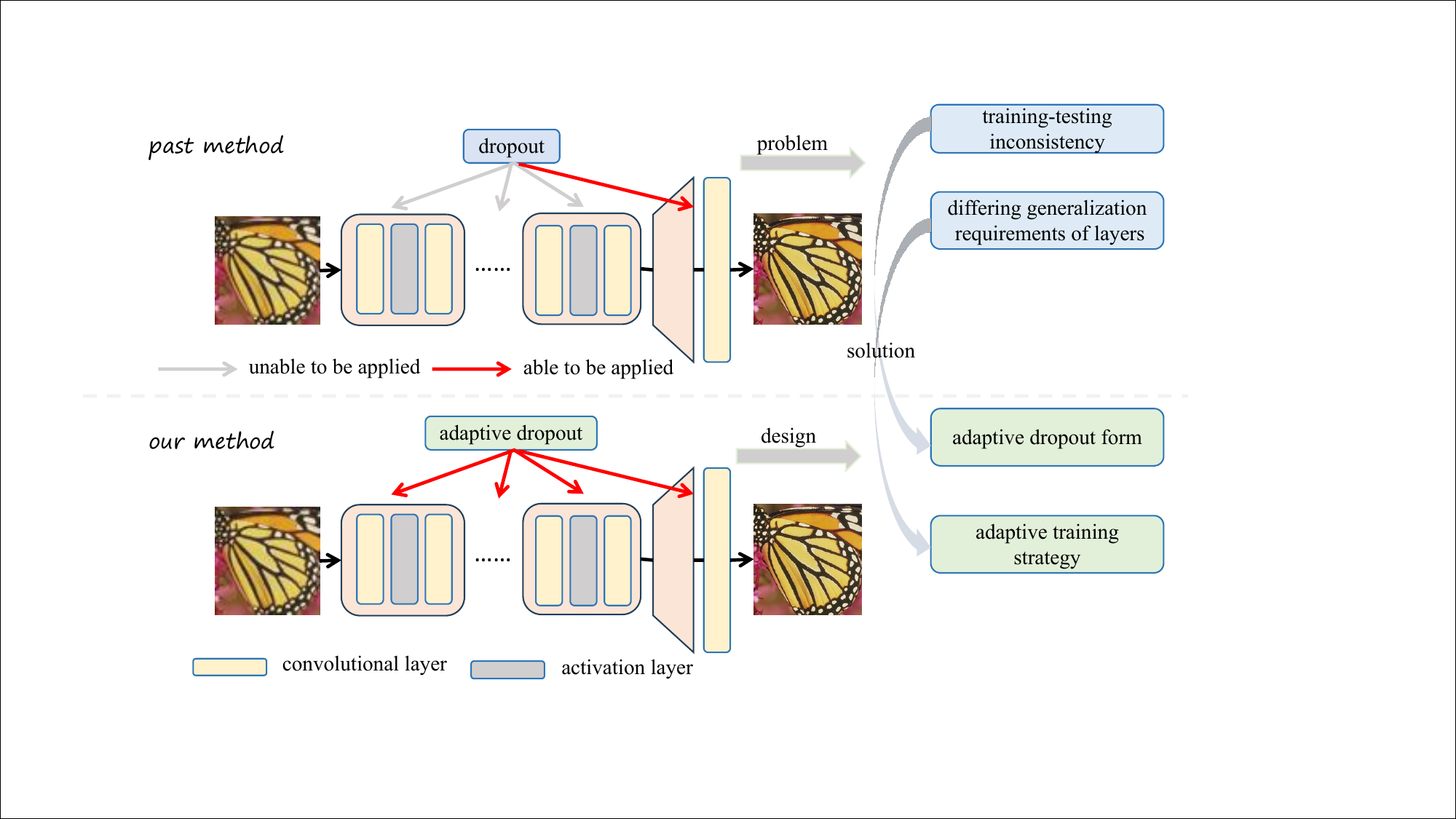}
\vspace{-5mm}
\caption{Placing dropout before the final convolutional layer can enhance the generalization performance of blind SR models, whereas placing it elsewhere has the opposite effect due to the inconsistency introduced. Our method effectively addresses the inconsistency allowing it to be easily applied to intermediate layers.}
\vspace{-5mm}
\label{fig:first}
\end{figure}
\section{Introduction}
\label{sec:intro}
Blind SR~\cite{gu2019blind, michaeli2013nonparametric, huang2020unfolding, liu2022blind, chen2022real} aims to recover high-quality images from low-quality images without knowing the degradation kernel. However, even with better real-world degradation simulation~\cite{wang2021real, sahak2023denoising, zhang2021designing} and more expressive model architectures~\cite{bell2019blind, liang2021flow, liang2021swinir}, blind SR models still encounter a severe over-fitting problem. Therefore, appropriate regularization terms~\cite{srivastava2014dropout, wu2021r, ghiasi2018dropblock} or training strategies~\cite{qian2022pointnext, bello2021revisiting} are urgently needed to enhance the generalization of blind SR models.
In the past, \citet{kong2022reflash} first introduced regularization into blind SR, adding dropout before the last convolutional layer of the model, which effectively improved the model's generalization ability. Subsequently, \citet{wang2024navigating} argued that dropout could harm the recovery of high-frequency image information and proposed Simple-Align, which feeds low-quality images obtained from different degradations of the same image into the network and aligns the features before the last convolutional layer, encouraging the model to be degradation-invariant.

It is worth noting that the above regularization methods all target the features before the last convolutional layer, while the features at intermediate layers are only implicitly impacted~\cite{neyshabur2017implicit, neyshabur2014search, sstkd_pami}. However, we find this indirect impact hard to regularize the features at intermediate layers to be degradation-invariant~\cite{huang2022learning, liu2022degradation, dlpl}, and instead exacerbates the imbalance between channels, indicating the necessity of explicit regularization for intermediate features. Therefore, we explore a new solution that regularizes the intermediate features on the shoulder of Dropout~\cite{kong2022reflash}, which further elevates the generalization ability of blind SR models. 

\begin{figure*}[t]
	\centering
	\includegraphics[width=1\linewidth]{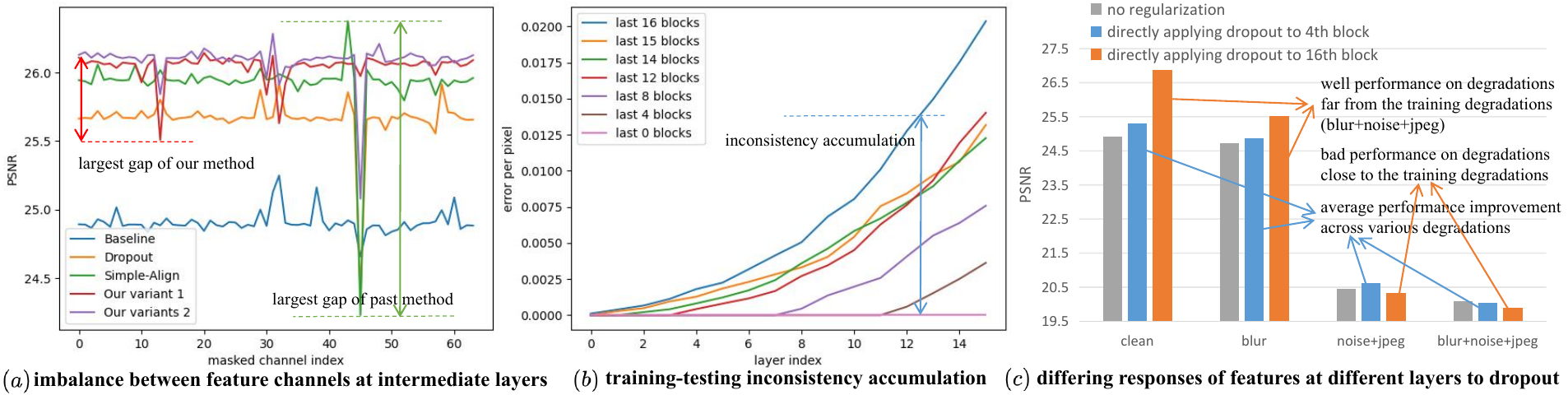}
    \vspace{-5mm}
	\caption{ \textbf{The source of our motivation.} (a) shows the imbalance between shallow feature channels, which is not well constrained by past methods. Mask channel index refers to the position of the channel we mask when adopting channel ablation. (b) compares the results of adding dropout to more blocks, indicating the accumulation of training-testing inconsistency along layers. (c) demonstrates the different behaviors of shallow and deep features when faced with dropout, as summarized in the text. We attribute this to the differing generalization requirements of different layers. Specifically, deep features more prominently represent information related to training degradations. Regularizing degradation representations alleviates the network's overfitting to training degradations, but directly applying dropout in deep layers also impairs the expressiveness of feature representations, leading to the performance shown in the chart. Shallow features, which contain more general representations, exhibit more consistent performance across wider degradations after being regularized.}
	\label{fig:motivation}
	\vspace{-5mm}
\end{figure*}

We rethink the dropout technique for this task. In previous discussions~\cite{ozgur2020effect, kong2022reflash}, it is not suitable for dropout to be directly applied in regression problems like SR. But~\citet{kong2022reflash} improved the model's generalization ability by adding dropout before the last convolutional layer, and indicated that adding it elsewhere led to performance drop. This raises the question of what causes this phenomenon. To analyze the reason, we conduct detailed experiments and attribute the performance decline to the inconsistency brought by directly applying dropout to intermediate layers. Specifically, as shown in~\figurename~\ref{fig:motivation}(b), directly applying dropout to intermediate layers introduces training-testing inconsistency, which harms the expressiveness of features considering the mean shift in the distribution fitted during training. Moreover, we observe that different layers of the network impact generalization differently, with shallow layers influencing general feature extraction and deep layers influencing degradation handling. Among them, general feature representation can consistently play a role in content reconstruction, while degradation representation needs regularization to perform well on unseen data. However, directly applying dropout to intermediate layers doesn't take into account the inconsistency in generalization requirements across different layers. These two types of inconsistency essentially require us to strike a balance between the fitting ability and generalization ability of the network.

In this paper, we introduce Adaptive Dropout for blind SR models to enhance generalization, consisting of the adaptive dropout format and the adaptive training strategy as shown in~\figurename~\ref{fig:model}. The key design of our method lies in its consideration of the balance between feature expressiveness and degradation generalization. Concretely, we redesign the format of dropout to adaptively integrate the dropouted feature with the original feature, which effectively mitigates training-testing inconsistency. Furthermore, considering the differing generalization requirements across different layers, we introduce the adaptive training strategy which adaptively helps blind SR models achieve a balance between fitting ability and generalization ability while satisfying the generalization requirements across different layers by layer-wise annealing. It is worth noting that, since our method is easy to integrate with Dropout~\cite{kong2022reflash}, we unify them and treat them as a whole to help blind SR models improve generalization. We show the superiority of our proposed method in \figurename~\ref{fig:first} and summarize our main contributions as follows:

\begin{itemize}[leftmargin=*]
    \item[$\bullet$] We specifically analyze the inconsistency introduced by directly applying dropout to intermediate layers, including training-testing inconsistency and the inconsistency in the generalization requirements across different layers.

    \item[$\bullet$] We introduce Adaptive Dropout that adaptively integrates the dropouted feature with the un-dropouted one. Moreover, we introduce an adaptive training strategy that achieves an elegant trade-off between fitting and generalizing across different layers by introducing different adaptive weight formats and layer-wise annealing.

    \item[$\bullet$] We validate the effectiveness of Adaptive Dropout on multiple synthetic datasets and real-world datasets. It is also highly effective in other image restoration tasks.
\end{itemize}
\section{Related Work}
\label{sec:related}

\subsection{Blind Super-Resolution}
\vspace{-1mm}
To deal with complicated degradation kernels~\cite{dong2022real} in the real world, blind SR~\cite{yue2022blind, qin2024multi} is introduced into the field of super-resolution, which aims to recover high-resolution images even though the degradations are unknown. Compared with implicit degradation modeling~\cite{zhang2022weakly, oh2023super, khan2024lightweight}, which requires large datasets during training, explicit degradation modeling~\cite{li2022learning, yan2024kgsr} performs better, where the degradations are added into images as data augmentation during training. DASR~\cite{liang2022efficient} introduces a new degradation scheme containing four types of degradation (bicubic, blur, noise, JPEG). Then in Real-ESRGAN~\cite{wang2021real}, applying the degradation scheme twice has proved helpful in generating low-resolution images closer to the real world, and has become a regular part of data augmentation when training blind SR models. However, just training blind SR models with a large degradation pool brings a severe problem: the SR network learns to overfit the distribution of specific degradations rather than the distribution of natural images~\cite{zhang2021designing, kong2022reflash, yue2022blind}. The problem forces researchers to find a proper training strategy to help blind SR models generalize better.

\subsection{Regularization methods for blind SR}
\vspace{-1mm}
For models of another field~\cite{urur, sstkd, pptformer}, regularization~\cite{srivastava2014dropout, wu2021r, ghiasi2018dropblock}, normalization~\cite{bjorck2018understanding, wu2018group, xu2019understanding}, and other modules~\cite{gpwformer, he2022masked, chen2023masked} are widely used as training strategies to improve the network's generalization performance. However, for blind SR models, few researchers pay attention to the training strategies. Previously, researchers focused more on improving the network's generalization ability to unknown degradations through complex degradation paradigms~\cite{wang2021real, sahak2023denoising, zhang2021designing} and network designs~\cite{bell2019blind, liang2021flow, liang2021swinir}, while overlooking the fact that training strategies are also an important factor affecting the network's generalization performance. Recently, ~\citet{kong2022reflash} demonstrated that dropout can also be applied to blind SR models and should be added before the last convolutional layer while adding dropout to other locations (such as within the residual blocks) leads to a decline in model performance. Simple-Align~\cite{wang2024navigating} shows the side-effect of dropout and provides a better choice as regularization, where two low-resolution images obtained from the same high-resolution image but with different degradations are fed into the network, and the statistics of the outputs before the last convolutional layer are aligned across different dimensions. However, we argue that dropout is still a great choice and can be added into all blocks in a different way.
\vspace{-2mm}

\section{Motivation}
\label{sec:motivation}
\subsection{Intermediate layers need explicit regularization}
\vspace{-1mm}
\label{sec:needs}

Both Dropout~\cite{kong2022reflash} and Simple-Align~\cite{wang2024navigating} regularize features before the final convolutional layer which directly impact the quality of restoration. So applying constraints to these features can effectively enhance the model's generalization ability. Meanwhile, regularizations targeting the final layer also serve as implicit constraints on intermediate layers. However, we argue this indirect constraint is insufficient to generalize the features at intermediate layers.

We evaluate the balance between channels in the intermediate layers under different regularization with channel ablation~\cite{morcos2018importance} and show the results in \figurename~\ref{fig:motivation}(a). Specifically, we sequentially occlude each channel to obtain a new feature map, which is then normalized to ensure the conservation of total energy, input into subsequent layers, and calculate the PSNR. Surprisingly, adding regularization only to the final layer doesn't improve the generalization of the features at intermediate layers, while stronger regularization exacerbates the imbalance between channels. In contrast, we explicitly regularize the features at intermediate layers resulting in a smaller performance decrease when conducting channel ablation~\cite{morcos2018importance}. This indicates the necessity of explicit regularization of the features at intermediate layers.

As a widely used regularization method, we want to incorporate dropout into intermediate layers of Blind SR networks to explicitly regularize features there. Although ~\citet{kong2022reflash} argue that dropout can only be applied before the last convolutional layer and not to intermediate layers, they lack an in-depth analysis of the reasons behind this. Therefore, we delve deeper by analyzing the issues brought about by dropout, seeking a reasonable approach to explicitly regularize features at intermediate layers.
\begin{figure*}[t]
	\centering
	\includegraphics[width=1\linewidth]{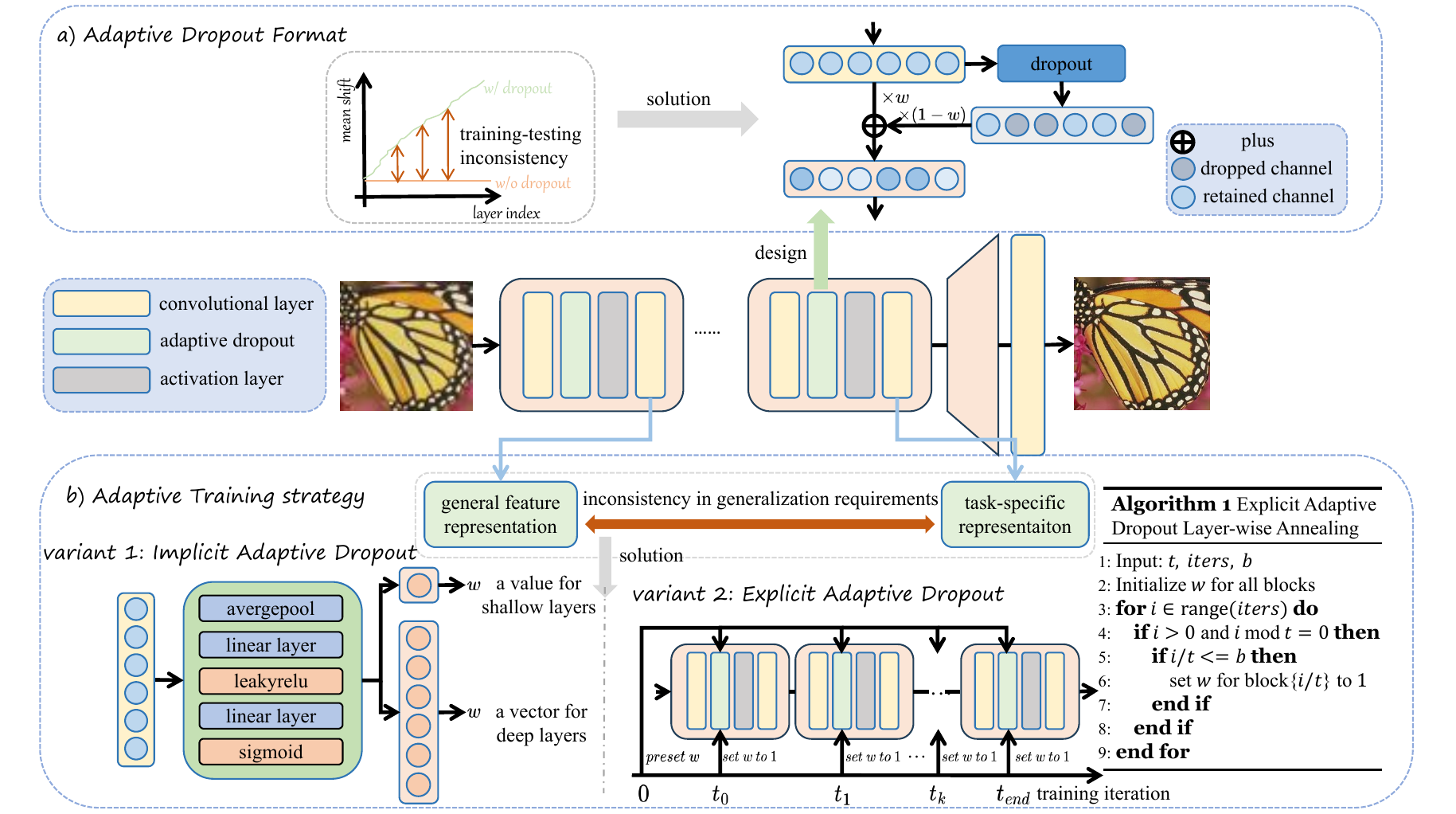}
    \vspace{-5mm}
	\caption{ \textbf{Overall schema of our proposed adaptive dropout.} Adaptive dropout consists of two components. (a) The adaptive dropout format adaptively couples the features before and after dropout, thereby achieving a balance between the network's fitting ability and generalization ability while alleviating the training-testing inconsistency. (b) Considering the different generalization requirements of features at different layers, we design an adaptive training strategy and its two variants. Regarding Implicit Adaptive Dropout, it adaptively adjusts the weight $w$ through a module, with different designs for shallow and deep layers. Regarding Explicit Adaptive Dropout, it initializes $w$ in all blocks and explicitly anneals it block-by-block during training, as shown in the algorithm beside.}
	\label{fig:model}
	\vspace{-5mm}
\end{figure*}
\subsection{Inconsistency when directly applying dropout}
\vspace{-1mm}
\label{sec:cannot}
In Dropout~\cite{kong2022reflash}, the impact of applying dropout to the network's intermediate layers was analyzed, leading to the following conclusions: (1) Applying dropout to the intermediate layers of the network rather than the final layer results in a performance decline. (2) Applying more dropout comes to more performance declines. (3) Applying dropout closer to the output layer results in better model performance. 

Overall, we attribute these three conclusions to the inconsistency brought by applying dropout to the intermediate layers, including that between the training and testing phases, and that among different layers of the network. Considering that the performance of Blind SR networks in unseen degradations and unseen scenarios is a joint effect of their fitting and generalization ability, improving generalization requires ensuring fitting ability. However, the inconsistency severely undermines the network's fitting ability. Below, we separately explain the causes of the inconsistency and provide corresponding solutions.

\noindent\textbf{Point 1: The training-testing inconsistency.} For conclusion (1) and (2) brought by Dropout~\cite{kong2022reflash}, We attribute them to the training-testing inconsistency~\cite{johnson2024inconsistency, li2019understanding} caused by applying dropout directly at intermediate layers. Moreover, when applying more dropout to intermediate layers, this inconsistency accumulates along the direction of information propagation, leading to the representations obtained from training not being well utilized during inference. In other words, the feature representations learned are losing their expressiveness. We illustrate the accumulation of inconsistency in~\ref{fig:motivation}, where the inconsistency is manifested as the shift in features at intermediate layers after applying dropout.

First, we introduce the training-testing inconsistency brought by dropout. We take channel dropout as an example. Consider a feature map $x_k \in {\mathbb{R}^{c \times h \times w}}$. When applying channel dropout to $x_k$, it is multiplied by a dropout mask $a_k \in {\mathbb{R}^c}$ which comes from a Bernoulli distribution during training: $\hat{x_k} = a_kx_k$. During training and testing, the mean of $\hat{x_k}$ remains the same, but there is a variance shift. Then after passing through the activation function, the variance shift leads to the mean shift which harms the expressiveness of the feature presentations. This issue does not exist when dropout is added before the last convolutional layer, with no activation function following it.
Assume we have $x_k$ with a mean $\mu$ and variance $\sigma^2$:
\begin{small}
\vspace{-1mm}
\begin{align*}
    \mathbf{var}[\hat{x_k}] = \begin{cases}
            \frac{1}{1-p}(\mu^2+\sigma^2)-\mu^2, &\textbf{during~~training} \\
            \sigma^2, &\textbf{during~~inference}
        \end{cases}
\end{align*}
\vspace{-1mm}
\end{small}
Then, the relative deviation of the variance of $\hat{x_k}$ during training can be expressed as $s_k$:
\begin{small}
    \begin{align*}
        s_k &= \frac{\mathbf{var}_\textit{training}[\hat{x_k}]}{\mathbf{var}_\textit{inference}[\hat{x_k}]}-1 \\
        &= \frac{1}{1-p}(\frac{\mu^2}{\sigma^2}+1)-\frac{\mu^2}{\sigma^2}-1 \\
        &= \frac{p}{1-p}(\frac{\mu^2}{\sigma^2}+1)
    \end{align*}
\end{small}
So reducing $s_k$ can decrease the training-testing inconsistency, while also reducing the inconsistency accumulation.

\noindent\textbf{Solution 1: Adaptive dropout format.} Integrating the dropouted feature with the original feature is a natural thought, considering we need to enhance the generalization of features while ensuring their expressiveness. Therefore, we introduce an adaptive dropout format to Blind SR, as shown in~\figurename~\ref{fig:model}(a). the basic format can expressed as:
\begin{equation}
\vspace{-1mm}
        f(x)=wx+(1-w)dropout(x,p)
\label{eq:basic}
\end{equation}
Where $x$ represents the feature map, $dropout$ refers to dropout operations, $w$ represents the weighted coefficients, and $p$ represents the dropout rate. We can calculate the relative deviation of the variance introduced by the adaptive dropout format in the same way:
\begin{small}
\begin{align*}
        & \mathbf{var}[x_k^{ad}] = \frac{{(1-w)}^2}{1-p}(\mu^2+\sigma^2)-\mu^2 \\
        & s_k^{ad} = \frac{{(1-w)}^2p}{1-p}(\frac{\mu^2}{\sigma^2}+1)
\end{align*}
\end{small}
Compared with standard dropout which can be viewed as a special case of adaptive dropout format when $w=0$, adaptive dropout format reduces the variance shift by ${(1-w)}^2$, leading to less training-testing inconsistency and inconsistency accumulation when applying it into multiple blocks.   

\begin{figure}[t]
\centering
\includegraphics[width=1\linewidth]{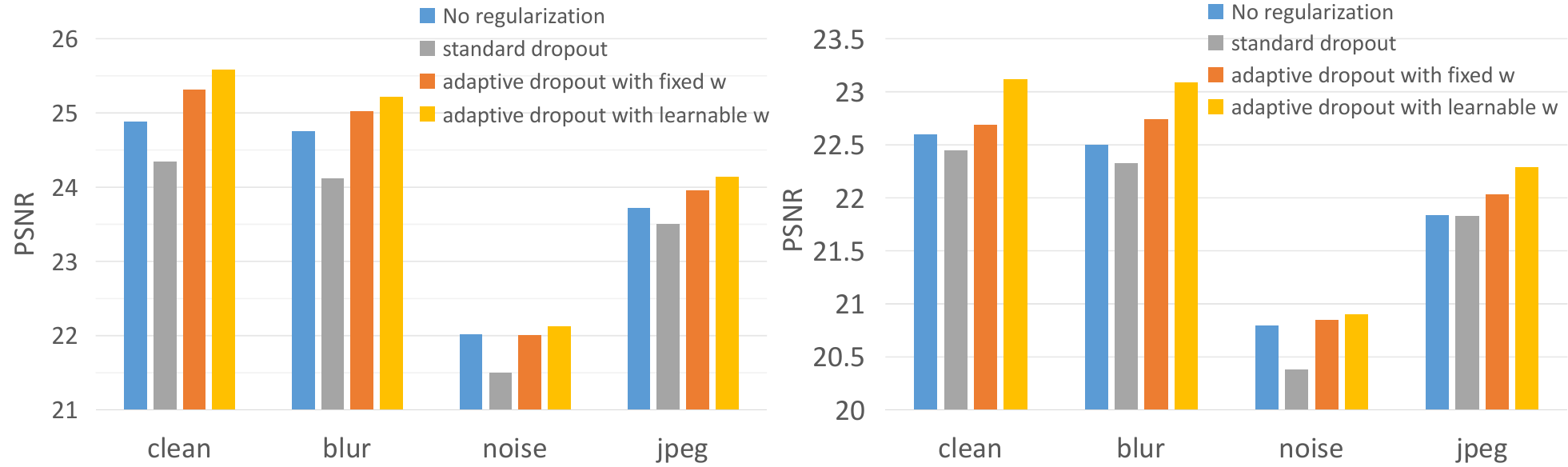}
\vspace{-5mm}
\caption{\textbf{Our exploration for mitigating the inconsistency in generalization requirements across different layers.} The left and right figures respectively show the model performance using different regularizations on Set5 and Set14. With consideration of how to better retain the fitting ability of SR models, a learnable weighted structure yields better results than the standard one.}
\label{fig:explore}
\vspace{-5mm}
\end{figure}

\noindent\textbf{Point 2: The inconsistency in generalization requirements across different layers.} For conclusion (3) brought by Dropout~\cite{kong2022reflash}, we attribute this to the differing generalization requirements of shallow and deep layers. Prior works~\cite{long2018transferable, yosinski2014transferable} have found that representations extracted from the shallow layers are more generalized, while the representations extracted from the deep layers show strong task relevance. For blind SR, the extraction of general features is crucial for image reconstruction, regardless of the degradation the low-quality image has undergone. However, task-specific representations may not perform well on test images, considering that these representations may overfit the degradations of the train set without regularization. Therefore, providing deeper features with greater perturbation without compromising fitting ability helps the model avoid overfitting to specific degradations in the training pool. To support this point, we verify the performance on Set5 with different degradations when dropout is directly applied at different layers, presenting the main results in \figurename~\ref{fig:motivation}(c). It can be observed that after being regularized, deep features result in differing performance across various degradations, whereas shallow features exhibit the opposite behavior, indicating their differing generalization requirements.

\noindent\textbf{Solution 2: Adaptive training strategy.} In the basic format of adaptive dropout, $x$ represents the complete transmission of information, $dropout(x)$ represents the regularizations on features and $w$ represents the trade-off between fitting ability and generalization ability. Considering the differing generalization requirements of different layers, it is reasonable to set different $w$ for each layer or block.  However, there is no clear guidance on how to apply different perturbations to different layers. So we design an experiment where we couple $x$ and $dropout(x)$ with learnable weights $w$ and $1-w$ for each block, to help uncover the patterns guiding the application of adaptive dropout format. Also, we simply use an adaptive dropout format with $w=0.5$ as a comparison. 

\begin{figure}[t]
\centering
\includegraphics[width=1\linewidth]{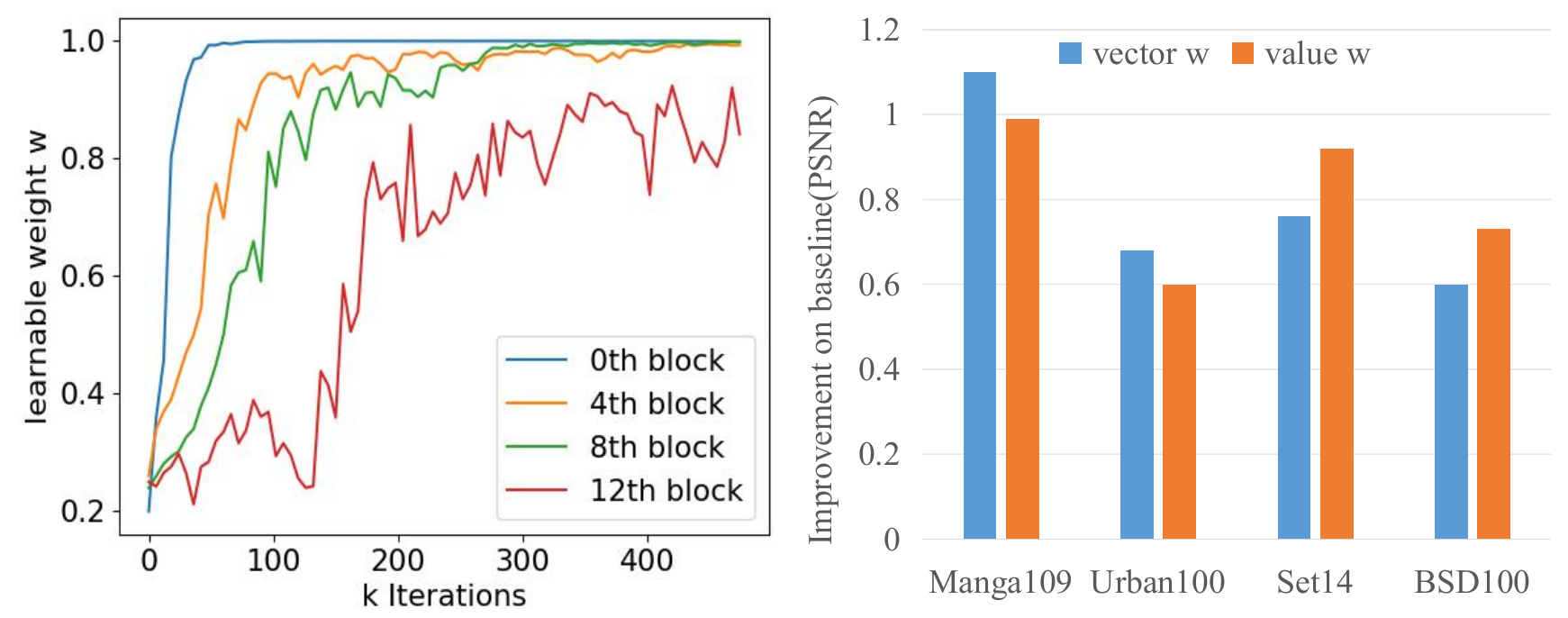}
\vspace{-5mm}
\caption{\textbf{Our exploration for the adaptive training strategy.} The graph on the left shows the trend of changes in $w$ across different layers when we use the learnable weight structure. The graph on the right illustrates the performance differences between \texttt{vector~w} and \texttt{value~w}, where \texttt{vector~w} performs better on more complex datasets while \texttt{value~w} performs better on others.}
\vspace{-5mm}
\label{fig:learnable_p}
\end{figure}

Understandably, this learnable weighted structure yields more significant effects than the simple application of the adaptive dropout format as illustrated in~\figurename~\ref{fig:explore}, for the former one ensures the network's fitting ability since the network adjusts $w$ towards fitting the train set. Then, we present the changes in $w$ across different blocks in~\figurename~\ref{fig:learnable_p}(a) to identify the patterns of the network's adaptive adjustment of $w$: First, $w$ in each block gradually approaches $1$, indicating that the model itself slowly reduces the perturbation of features, which represents the requirements of SR network for fitting ability. Second, compared to deeper blocks, the $w$ values in shallower blocks approach $1$ more rapidly, representing the requirements for generalization ability across different layers. We can summarize that the network is performing an adaptive annealing process from shallow to deep layers. This layer-wise annealing training strategy, which we refer to as the adaptive training strategy shown in~\figurename~\ref{fig:model}(b), aligns with our requirement to address the inconsistency in generalization needs across different layers. We divide the training phase into generalization and fitting stages, referring to the periods when intermediate layer features are perturbed and not perturbed, respectively. The shallow layers of the network enter the fitting stage earlier, ensuring that the model's general feature representations have good expressiveness. The deep layers remain in the generalization stage for a longer time, ensuring that the model does not overfit the degradations of the train set, thereby performing well on unseen degradations. Additionally, the adaptive training strategy effectively addresses the inconsistency accumulation issue, enabling the network to eliminate perturbations from the source.

\section{Methods}
\label{sec:variants}

To further improve the generalization of blind SR networks, we propose adaptive dropout, where the adaptive dropout format and adaptive training strategy are the core designs of our method. Below, we separately introduce them in detail.

\subsection{Adaptive Dropout format}
\vspace{-1mm}
\label{sec:form}
In Equation~\ref{eq:basic}, we presented the basic form of the adaptive dropout format, so here we specifically discuss the hyper-parameters, Integration, and application in networks.

\noindent\textbf{Hyper-parameters.} The adaptive dropout format has two hyper-parameters, $w$ and $p$. However, we find that setting $p$ to $0.5$ and adjusting $w$ alone is sufficient to achieve better generalization performance in blind SR networks. Therefore, we simply set $p$ to $0.5$ to simplify our method design. 

\noindent\textbf{Integration.} Since the proposed adaptive dropout targets the intermediate features of the network, it can effectively be integrated with Dropout~\cite{kong2022reflash} which targets features before the last convolutional layer. We still retain the original format of dropout before the last convolutional layer for it doesn't bring the training-testing inconsistency.

\noindent\textbf{Application.} In CNN-based SR networks~\cite{ledig2017photo, zhang2018image}, using multiple adaptive dropout within the same block does not improve performance on unseen degradation. We speculate that this might excessively constrain the network's fitting capability within the same training time. For SwinIR~\cite{liang2021swinir} where dropout is generally not used under normal settings, we reintroduce this regularization method and replace all dropout with adaptive dropout, which is also applicable to other transformer-based SR networks~\cite{hsu2024drct, chen2023activating}.

\subsection{Adaptive Training Strategy}
\vspace{-1mm}
\label{sec:strategy}
Previously, we discovered that the network implicitly employs a layer-wise annealing strategy through our attempt to adaptively adjust the learnable $w$. Given this, we argue that explicitly applying this strategy can also yield good results and design two variants for adaptive dropout.

\noindent\textbf{Implicit Adaptive Dropout with different formats of $w$.} We implicitly adopt the adaptive training strategy by treating $w$ as an optimizable variable and letting the neural network learn the optimal $w$ just as our attempt to adaptively adjust the learnable $w$. To better weigh the importance of each channel, we use a network similar to channel attention~\cite{hu2018squeeze}, where the weight $w_i$ can be obtained by:
\begin{small}
\begin{align*}
        &w_i = sigmoid(f_2(ReLU(f_1(x_k)))) \\
        &\hat{x_k} = wx_k + (1-w)dropout(x_k)
\end{align*}
\end{small}
Where $f_1$ and $f_2$ represent linear projection, $x_k$  means the original feature, and $\hat{x_k}$ means the coupled feature.
According to specific requirements, $w_i$ can take the format of $[B, C]$, where $B$ means the batch size and $C$ means the number of channels, meaning that different weights are learned for each channel of the feature, referred to as \texttt{vector~w}. It can also take the format of $[B, 1]$, meaning that the same weight is shared across all channels, referred to as \texttt{value~w}. The former introduces more randomness and variability compared to the latter. The impact of this randomness is shown in~\figurename~\ref{fig:learnable_p}(b): in more complex datasets, Manga109 and Urban100, \texttt{vector~w} achieves better results than \texttt{value~w}, whereas in simpler datasets, Set14 and BSD100, the opposite is true. Therefore, we use them in different blocks to strike a balance between the two. Considering that we should reduce disturbances in the shallow layers, we apply \texttt{value~w} in the shallow layers and \texttt{vector~w} in the deeper layers.

\noindent\textbf{Explicit Adaptive Dropout with layer-wise annealing.} We explicitly adopt the adaptive training strategy by initializing $w$ for all blocks and then annealing it block-by-block during training. Specifically, for every $t$ training iteration, we set the $w$ to $0$ in one block, repeating the operation from shallow to deep layers, where $t$ is related to the total number of training iterations. In this way, general feature representations can be sufficiently preserved, while degradation-related representations can be sufficiently constrained.

\begin{figure*}[ht]
\centering
\includegraphics[width=1\linewidth]{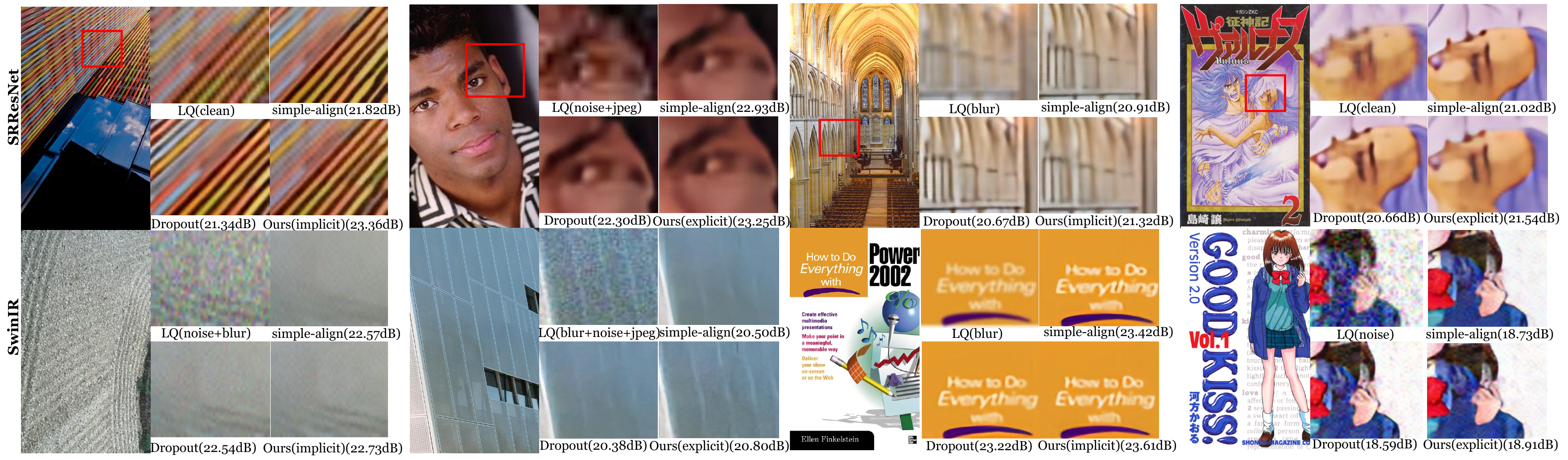}
\vspace{-7mm}
\caption{\textbf{Comparison with other regularization methods~\cite{kong2022reflash, wang2024navigating}.}}
\vspace{-1mm}
\label{fig:comparison}
\end{figure*}

\begin{table*}[t]
  \small
  \setlength{\abovecaptionskip}{-2pt}
	\setlength{\belowcaptionskip}{-4pt}
	\renewcommand{\arraystretch}{0.9}
  \begin{center}
    \caption{\textbf{The PSNR ($\mathbf{dB}$) results of models with $\times 4$.} We test them on eight types of degradations or multi-degradations ($b$ is $blur$, $n$ is $noise$, $j$ is $jpeg$). \textbf{\textcolor{red}{Red}} and \underline{\textcolor{blue}{Blue}} indicate the best and the second-best performance, respectively.}
      \label{tab:comparison}
  \resizebox{0.85\linewidth}{!}{
    \begin{tabular}{|l|c|cc|cc|cc|cc|cc|}
    \hline
        \multirow{2}{*}{Models} &\multirow{2}{*}{Regularization}  & \multicolumn{2}{c|}{Set5~\cite{bevilacqua2012low}} & \multicolumn{2}{c|}{Set14~\cite{yang2010image}} & \multicolumn{2}{c|}{BSD100~\cite{martin2001database}} & \multicolumn{2}{c|}{Manga109~\cite{matsui2017sketch}} &  \multicolumn{2}{c|}{Urban100~\cite{huang2015single}} \\ \cline{3-12}
        &  & \multicolumn{1}{c}{clean} & blur  & \multicolumn{1}{c}{clean} & blur  & \multicolumn{1}{c}{clean} & blur  & \multicolumn{1}{c}{clean} & blur  & \multicolumn{1}{c}{clean} & blur \\ \hline
        \multirow{4}{*}{SRResNet~\cite{ledig2017photo}} & None    & 24.89  & 24.76  & 22.60  & 22.50  & 23.06  & 22.99  & 18.42  & 18.75  & 21.24  & 21.06  \\ 
        ~ & Dropout~\cite{kong2022reflash}                   & 25.67  & 25.34  & 23.11  & 22.92  & 23.37  & 23.24  & 18.87  & 19.14  & 21.58  & 21.31  \\ 
        ~ & Explicit Adaptive Dropout & \textbf{\textcolor{red}{26.11}} & \underline{\textcolor{blue}{25.39}} & \underline{\textcolor{blue}{23.42}} & \underline{\textcolor{blue}{23.20}} & \textbf{\textcolor{red}{23.76}} & \textbf{\textcolor{red}{23.59}} & \underline{\textcolor{blue}{19.40}} & \underline{\textcolor{blue}{19.62}} & \underline{\textcolor{blue}{21.87}} & \textbf{\textcolor{red}{21.48}} \\ 
        ~ & Implicit Adaptive Dropout & \underline{\textcolor{blue}{26.09}} & \textbf{\textcolor{red}{25.56}} & \textbf{\textcolor{red}{23.47}} & \textbf{\textcolor{red}{23.21}} & \underline{\textcolor{blue}{23.73}} & \underline{\textcolor{blue}{23.50}} & \textbf{\textcolor{red}{19.51}} & \textbf{\textcolor{red}{19.69}} & \textbf{\textcolor{red}{21.90}} & \underline{\textcolor{blue}{21.40}} \\ \hline
        \multirow{4}{*}{SwinIR~\cite{liang2021swinir}} & None      & 26.25  & 26.03  & 23.76  & \underline{\textcolor{blue}{23.47}}  & 23.91  & \underline{\textcolor{blue}{23.83}}  & 19.10  & 19.19  & 22.18  & 21.90   \\ 
        ~ & Dropout~\cite{kong2022reflash}                   & 26.32  & 26.08  & 23.80  & 23.36  & 23.90  & \textbf{\textcolor{red}{23.87}}  & 19.15  & 19.30  & 22.27  & 21.99  \\ 
        ~ & Explicit Adaptive Dropout & \textbf{\textcolor{red}{26.54}} & \textbf{\textcolor{red}{26.21}} & \underline{\textcolor{blue}{23.92}} & \textbf{\textcolor{red}{23.56}} & \textbf{\textcolor{red}{24.09}} & \textcolor{black}{23.59} & \underline{\textcolor{blue}{19.22}} & \underline{\textcolor{blue}{19.39}} & \textbf{\textcolor{red}{22.41}} & \textbf{\textcolor{red}{22.19}} \\ 
        ~ & Implicit Adaptive Dropout & \underline{\textcolor{blue}{26.46}} & \underline{\textcolor{blue}{26.12}} & \textbf{\textcolor{red}{23.95}} & \textcolor{black}{23.41} & \underline{\textcolor{blue}{24.01}} & \textcolor{black}{23.50} & \textbf{\textcolor{red}{19.31}} & \textbf{\textcolor{red}{19.44}} & \underline{\textcolor{blue}{22.32}} & \underline{\textcolor{blue}{22.16}} \\ \hline
        & &  \multicolumn{1}{c}{noise} & jpeg & \multicolumn{1}{c}{noise} & jpeg & \multicolumn{1}{c}{noise} & jpeg & \multicolumn{1}{c}{noise} & jpeg & \multicolumn{1}{c}{noise} & jpeg \\ \hline
        \multirow{4}{*}{SRResNet~\cite{ledig2017photo}} & None    & 22.02  & 23.72  & 20.81  & 21.84  & 20.34  & 22.48  & 19.74  & 18.30  & 19.73  & 20.60  \\ 
        ~ & Dropout~\cite{kong2022reflash}                   & 21.87  & 24.04  & 20.61  & 22.16  & 21.00  & 22.70  & 18.25  & 18.62  & 19.58  & 20.89  \\ 
        ~ & Explicit Adaptive Dropout & \textbf{\textcolor{red}{22.41}} & \textbf{\textcolor{red}{24.38}} & \underline{\textcolor{blue}{21.14}} & \underline{\textcolor{blue}{22.46}} & \underline{\textcolor{blue}{21.48}} & \textbf{\textcolor{red}{23.00}} & \underline{\textcolor{blue}{18.67}} & \underline{\textcolor{blue}{19.07}} & \textbf{\textcolor{red}{19.90}} & \underline{\textcolor{blue}{21.11}} \\ 
       ~ & Implicit Adaptive Dropout & \textbf{\textcolor{red}{22.41}} & \underline{\textcolor{blue}{24.36}} & \textbf{\textcolor{red}{21.15}} & \textbf{\textcolor{red}{22.51}} & \textbf{\textcolor{red}{21.53}} & \textbf{\textcolor{red}{23.00}} & \textbf{\textcolor{red}{18.76}} & \textbf{\textcolor{red}{19.20}} & \textbf{\textcolor{red}{19.90}} & \textbf{\textcolor{red}{21.15}} \\ \hline
        \multirow{4}{*}{SwinIR~\cite{liang2021swinir}} & None      & 22.96  & 24.37  & 21.56  & 23.04  & 22.12  & 23.04  & 18.71  & 18.95  & 20.56  & 21.32  \\ 
        ~ & Dropout~\cite{kong2022reflash}                   & 23.12  & 24.41  & 21.59  & 23.09  & 22.10  & 23.08  & 18.73  & 19.03  & 20.67  & 21.38  \\ 
        ~ & Explicit Adaptive Dropout & \textbf{\textcolor{red}{23.63}} & \textbf{\textcolor{red}{24.68}} & \underline{\textcolor{blue}{21.67}} & \textbf{\textcolor{red}{23.29}} & \textbf{\textcolor{red}{22.24}} & \textbf{\textcolor{red}{23.29}} & \underline{\textcolor{blue}{18.92}} & \underline{\textcolor{blue}{19.27}} & \textbf{\textcolor{red}{20.85}} & \textbf{\textcolor{red}{21.52}} \\ 
        ~ & Implicit Adaptive Dropout & \underline{\textcolor{blue}{23.49}} & \underline{\textcolor{blue}{24.49}} & \textbf{\textcolor{red}{21.69}} & \underline{\textcolor{blue}{23.19}} & \underline{\textcolor{blue}{22.19}} & \underline{\textcolor{blue}{23.28}} & \textbf{\textcolor{red}{19.02}} & \textbf{\textcolor{red}{19.31}} & \underline{\textcolor{blue}{20.74}} & \underline{\textcolor{blue}{21.40}} \\ \hline
        ~ & ~ & b+n & b+j & b+n & b+j & b+n & b+j & b+n & b+j & b+n & b+j \\ \hline
        \multirow{4}{*}{SRResNet~\cite{ledig2017photo}} & None    & 23.31  & 23.44  & 21.81  & 21.70  & 22.27  & 22.34  & 18.60  & 18.53  & 20.46  & 20.30  \\ 
        ~ & Dropout~\cite{kong2022reflash}                   & 23.51  & 23.67  & 21.87  & 22.01  & 22.24  & 22.54  & 18.94  & 18.81  & 20.48  & 20.53  \\ 
        ~ & Explicit Adaptive Dropout & \textbf{\textcolor{red}{23.74}} & \textbf{\textcolor{red}{23.80}} & \textbf{\textcolor{red}{22.15}} & \textbf{\textcolor{red}{22.29}} & \textbf{\textcolor{red}{22.48}} & \textbf{\textcolor{red}{22.81}} & \underline{\textcolor{blue}{19.26}} & \underline{\textcolor{blue}{19.23}} & \textbf{\textcolor{red}{20.57}} & \textbf{\textcolor{red}{20.66}} \\ 
        ~ & Implicit Adaptive Dropout & \underline{\textcolor{blue}{23.65}} & \underline{\textcolor{blue}{23.73}} & \underline{\textcolor{blue}{22.12}} & \textbf{\textcolor{red}{22.29}} & \underline{\textcolor{blue}{22.44}} & \underline{\textcolor{blue}{22.78}} & \textbf{\textcolor{red}{19.51}} & \textbf{\textcolor{red}{19.29}} & \underline{\textcolor{blue}{20.50}} & \underline{\textcolor{blue}{20.63}} \\ \hline
        \multirow{4}{*}{SwinIR~\cite{liang2021swinir}} & None      & 23.80  & 23.84  & 22.20  & 22.26  & 22.61  & 22.82  & 19.07  & 19.02  & 20.89  & 20.79  \\ 
        ~ & Dropout~\cite{kong2022reflash}                   & 24.00  & 23.93  & 22.40  & 22.24  & 22.68  & 22.80  & 19.12  & 18.98  & \underline{\textcolor{blue}{20.92}}  & 20.91  \\ 
        ~ & Explicit Adaptive Dropout & \textbf{\textcolor{red}{24.24}} & \underline{\textcolor{blue}{24.27}} & \textbf{\textcolor{red}{22.91}} & \textbf{\textcolor{red}{22.41}} & \textbf{\textcolor{red}{22.81}} & \textbf{\textcolor{red}{23.05}} & \underline{\textcolor{blue}{19.25}} & \underline{\textcolor{blue}{19.29}} & \textbf{\textcolor{red}{21.00}} & \textbf{\textcolor{red}{20.96}} \\ 
        ~ & Implicit Adaptive Dropout & \underline{\textcolor{blue}{24.18}} & \textbf{\textcolor{red}{24.30}} & \underline{\textcolor{blue}{22.78}} & \underline{\textcolor{blue}{22.36}} & \underline{\textcolor{blue}{22.80}} & \textbf{\textcolor{red}{23.05}} & \textbf{\textcolor{red}{19.31}} & \textbf{\textcolor{red}{19.37}} & \underline{\textcolor{blue}{20.92}} & \underline{\textcolor{blue}{20.94}} \\ \hline
        ~ & ~ & n+j & b+n+j & n+j & b+n+j & n+j & b+n+j & n+j & b+n+j & n+j & b+n+j \\ \hline
        \multirow{4}{*}{SRResNet~\cite{ledig2017photo}} & None    & 23.21  & 22.70  & 21.59  & 21.44  & 22.24  & 22.05  & 18.25  & 18.43  & 20.42  & 20.10  \\ 
        ~ & Dropout~\cite{kong2022reflash}                   & 23.55  & 22.99  & 21.86  & 21.64  & 22.39  & 22.17  & 18.57  & 18.71  & 20.64  & 20.23  \\ 
        ~ & Explicit Adaptive Dropout & \underline{\textcolor{blue}{23.71}} & \textbf{\textcolor{red}{23.10}} & \underline{\textcolor{blue}{22.09}} & \textbf{\textcolor{red}{21.84}} & \underline{\textcolor{blue}{22.59}} & \underline{\textcolor{blue}{22.31}} & \underline{\textcolor{blue}{18.95}} & \underline{\textcolor{blue}{19.04}} & \underline{\textcolor{blue}{20.79}} & \textbf{\textcolor{red}{20.29}} \\
        ~ & Implicit Adaptive Dropout & \textbf{\textcolor{red}{23.74}} & \textbf{\textcolor{red}{23.10}} & \textbf{\textcolor{red}{22.13}} & \textbf{\textcolor{red}{21.84}} & \textbf{\textcolor{red}{22.60}} & \textbf{\textcolor{red}{22.32}} & \textbf{\textcolor{red}{19.07}} & \textbf{\textcolor{red}{19.10}} & \textbf{\textcolor{red}{20.81}} & \underline{\textcolor{blue}{20.27}} \\ \hline
        \multirow{4}{*}{SwinIR~\cite{liang2021swinir}} & None      & 23.67  & 22.99  & 22.11  & 21.82  & 22.61  & 22.34  & 18.79  & 18.80  & 20.98  & 20.45  \\ 
        ~ & Dropout~\cite{kong2022reflash}                   & 23.65  & 23.09  & 22.17  & 21.84  &  \underline{\textcolor{blue}{22.64}}  & 22.33  & 18.75  & 18.84  & \underline{\textcolor{blue}{21.12}}  & 20.55  \\ 
        ~ & Explicit Adaptive Dropout & \textbf{\textcolor{red}{23.82}} & \textbf{\textcolor{red}{23.18}} & \underline{\textcolor{blue}{22.30}} & \underline{\textcolor{blue}{22.19}} & \textbf{\textcolor{red}{22.78}} & \textbf{\textcolor{red}{22.50}} & \underline{\textcolor{blue}{18.82}} & \underline{\textcolor{blue}{18.91}} & \textbf{\textcolor{red}{21.15}} & \textbf{\textcolor{red}{20.57}} \\
       ~ & Implicit Adaptive Dropout & \underline{\textcolor{blue}{23.76}} & \underline{\textcolor{blue}{23.13}} & \textbf{\textcolor{red}{22.23}}  & \textbf{\textcolor{red}{22.19}} & \textcolor{black}{22.57} & \underline{\textcolor{blue}{22.39}} & \textbf{\textcolor{red}{18.87}} & \textbf{\textcolor{red}{19.10}} & \textcolor{black}{21.02} & \underline{\textcolor{blue}{20.56}} \\ \hline
    \end{tabular}}
  \end{center}

  \vskip -0.30cm
  \end{table*}

\begin{table*}[t]
    \small
  \setlength{\abovecaptionskip}{-2pt}
	\setlength{\belowcaptionskip}{-4pt}
	\renewcommand{\arraystretch}{0.9}
 \begin{center}
  \caption{\textbf{Comparison with Simple-Align~\cite{wang2024navigating} on synthetic datasets and real-world datasets.}}
  \label{tab:comparison2}
  \resizebox{0.85\linewidth}{!}{
    \begin{tabular}{|l|ccccc|ccccc|}
    \hline
         \multirow{2}{*}{Models+Regularization} & \multicolumn{5}{c|}{Synthetic dataset} &\multicolumn{5}{c|}{Real-world dataset} \\ \cline{2-11}
                &Set5 &Set14 &BSD100 &Manga109 &Urban100  &RealSR &DRealSR &mild &difficult & wild\\ \hline
         SRResNet+Simple-Align~\cite{wang2024navigating} & \underline{\textcolor{blue}{24.05}} & 22.31 & 22.70 & 19.11 & 20.80 & 24.63 & 27.14 & 17.19 & 18.05 & 17.81 \\
SRResNet+Explicit Adaptive Dropout & \textbf{\textcolor{red}{24.10}} & \textbf{\textcolor{red}{22.34}} & \underline{\textcolor{blue}{22.73}} & \underline{\textcolor{blue}{19.15}} & \underline{\textcolor{blue}{20.81}} & \underline{\textcolor{blue}{25.07}} & \textbf{\textcolor{red}{27.51}} & \underline{\textcolor{blue}{17.29}} & \textbf{\textcolor{red}{18.18}} & \underline{\textcolor{blue}{17.91}} \\
SRResNet+Implicit Adaptive Dropout & \underline{\textcolor{blue}{24.05}} & \underline{\textcolor{blue}{22.32}} & \textbf{\textcolor{red}{22.75}} & \textbf{\textcolor{red}{19.24}} & \textbf{\textcolor{red}{20.84}} & \textbf{\textcolor{red}{25.16}} & \textbf{\textcolor{red}{27.51}} & \textbf{\textcolor{red}{17.32}} & \textbf{\textcolor{red}{18.18}} & \textbf{\textcolor{red}{17.92}} \\
\hline
         RRDB+Simple-Align~\cite{wang2024navigating} & 24.33 & \textcolor{black}{22.51} & \textcolor{black}{22.90} & 19.14 & \textbf{\textcolor{red}{21.05}} & 24.76 & 27.00 & 17.14 & 18.08 & 17.81 \\
RRDB+Explicit Adaptive Dropout & \textbf{\textcolor{red}{24.40}} & \underline{\textcolor{blue}{22.56}} & \textbf{\textcolor{red}{22.97}} & \underline{\textcolor{blue}{19.27}} & \underline{\textcolor{blue}{21.00}} & \textbf{\textcolor{red}{25.39}} & \underline{\textcolor{blue}{27.46}} & \textbf{\textcolor{red}{17.62}} & \textbf{\textcolor{red}{18.33}} & \textbf{\textcolor{red}{18.04}} \\
RRDB+Implicit Adaptive Dropout & \underline{\textcolor{blue}{24.38}} & \textbf{\textcolor{red}{22.61}} & \underline{\textcolor{blue}{22.91}} & \textbf{\textcolor{red}{19.30}} & 20.96 & \underline{\textcolor{blue}{25.35}} & \textbf{\textcolor{red}{27.53}} & \underline{\textcolor{blue}{17.27}} & \underline{\textcolor{blue}{18.20}} & \underline{\textcolor{blue}{17.90}} \\
\hline
    \end{tabular}}
    \end{center}
    \vskip -0.30cm
\end{table*}

\begin{figure}[h]
\centering
\includegraphics[width=1\linewidth]{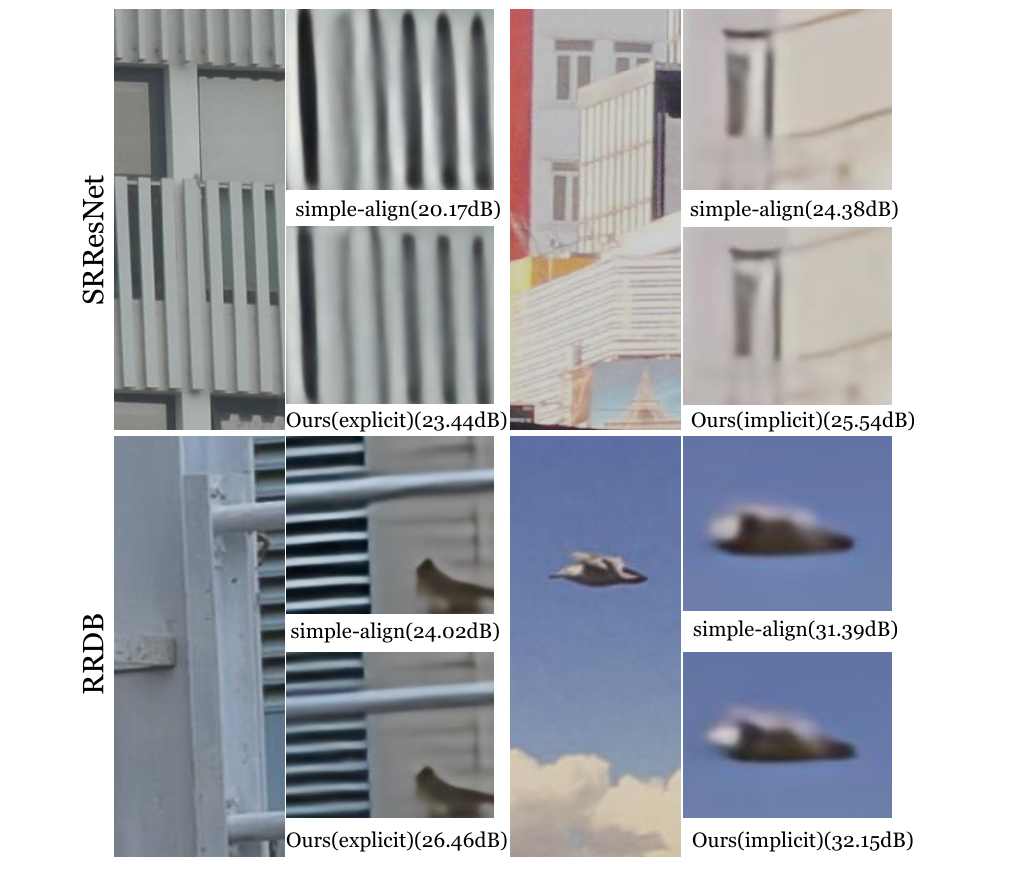}
\vspace{-5mm}
\caption{\textbf{Comparison with Simple-Align~\cite{wang2024navigating} on real-world datasets.}}
\label{fig:comparison2}
\vspace{-5mm}
\end{figure}
\section{Experiments}
\label{sec:experiments}
\noindent\textbf{Experimental settings.} We follow the multi-degradation settings and experimental configurations outlined in Dropout~\cite{kong2022reflash}. For training, we utilize the HR images from the DIV2K dataset~\cite{Agustsson_2017_CVPR_Workshops}. During training, we employ L1 loss and the Adam optimizer~\cite{diederik2014adam}. The batch size is set to 16, and the size of the LR images is $32\times32$. We adopt a cosine annealing strategy to dynamically adjust the learning rate, with an initial learning rate set to $2\times10^{-4}$. We train for a total of 500,000 iterations. For testing, we use the synthetic datasets Set5~\cite{bevilacqua2012low}, Set14~\cite{yang2010image}, BSD100~\cite{martin2001database}, Manga109~\cite{matsui2017sketch}, and Urban100~\cite{huang2015single}, as well as the real-world datasets NTIRE 2018 SR challenge data~\cite{timofte2017ntire}, RealSR~\cite{cai2019toward}, and DrealSR~\cite{wei2020component}. We compare our approach with Dropout~\cite{kong2022reflash} and Simple-Align~\cite{wang2024navigating}.

\noindent\textbf{Comparison with baseline models.} We present the experimental results for the baseline with different regularization in \tablename~\ref{tab:comparison}. It can be observed that our proposed adaptive dropout significantly outperforms the baseline and comparison methods across all degradation settings in all datasets. Based on the experimental results, we can draw the following conclusions: (1) Both Implicit Adaptive Dropout and Explicit Adaptive Dropout can bring more performance increase compared with Dropout~\cite{kong2022reflash}, for both of these variants explicitly regularize features at intermediate layers, encouraging models to learn more general features representations. (2) Explicit Adaptive Dropout generally outperforms Implicit Adaptive Dropout, considering that Implicit Adaptive Dropout focuses more on maintaining the model's fitting ability. We show the comparison results on RealSRResNet~\cite{ledig2017photo} and SwinIR~\cite{liang2021swinir} in \figurename~\ref{fig:comparison}.

\noindent\textbf{Comparison with Simple-Align.} The comparison results between our method and Simple-Align~\cite{wang2024navigating} are also shown in \figurename~\ref{fig:comparison}. Our proposed method outperforms Simple-Align~\cite{wang2024navigating} on all synthetic datasets, with notable improvements on the more complex datasets Manga109 and Urban100, showing better generalization ability on complex scenarios. \textit{The detailed data is in supplementary material.}

To further demonstrate the effectiveness of our method, we compare it with previous methods on real-world datasets, for low-resolution images collected from the real world are far more challenging to reconstruct than synthetic low-resolution images. RealSR~\cite{cai2019toward} and DRealSR~\cite{wei2020component} are the latest two paired datasets for real-world super-resolution, and we show the comparison results of these two datasets in \figurename~\ref{fig:comparison2} and \tablename~\ref{tab:comparison2}. We also conduct tests on the realistic NTIRE 2018 SR challenge data~\cite{timofte2017ntire}, and the results are presented in \tablename~\ref{tab:comparison2}. Our method outperforms Simple-Align~\cite{wang2024navigating} on all five real-world datasets.


\noindent\textbf{Applications in single-degradation task.} Adaptive Dropout also works well for generalization tasks of single degradation scenarios. We adopt our proposed methods on image denoising, image draining, and image dehazing. For image deraining, we train RCDNet~\cite{wang2023rcdnet} and Restormer~\cite{zamir2022restormer} on Rain100L~\cite{jiang2020multi} and test them on Rain100H~\cite{jiang2020multi}. For image denoising, we follow ~\citet{chen2023masked}, and train SwinIR on Gaussian noise. For image dehazing, we train FFANet~\cite{qin2020ffa} on ITS Dataset~\cite{li2019benchmarking} and test it on OTS Dataset~\cite{li2019benchmarking}. Results in \figurename~\ref{tab:single_task} illustrate that our methods help models generalize better on unseen degradation.

\noindent\textbf{Applications in generative models.} We show the results on generative IR in Tab.~\ref{tab:gen_IR}, where our methods work well on both GAN-based and diffusion-based blind SR models.

\begin{table}[h]
    \centering
    \small
  \setlength{\abovecaptionskip}{-0pt}
	\setlength{\belowcaptionskip}{-4pt}
	\renewcommand{\arraystretch}{0.9}
     \caption{\textbf{The performance of the model in single degradation tasks when augmented with explicit adaptive dropout.}}
     \label{tab:single_task}
    \begin{tabular}{ccccccc}
    \hline
         task &\multicolumn{2}{c}{models}  &PSNR &SSIM \\ \hline
         \multirow{4}{*}{derain} &\multirow{2}{*}{RCDNet} &None &17.36 &0.554 \\
         ~ &  &w/ ours &\textbf{17.59} &\textbf{0.565} \\ \cline{2-5}
         ~ &\multirow{2}{*}{Restormer} &None &29.46 &0.887 \\
         ~ & &w/ ours &\textbf{29.79} &\textbf{0.898} \\ \hline
         \multirow{2}{*}{dehaze} &\multirow{2}{*}{FFANet} &None &20.26 &0.881 \\
         ~ & &w/ ours &\textbf{21.09} &\textbf{0.883} \\ \hline
         \multirow{2}{*}{denoise} &\multirow{2}{*}{SwinIR} &None &26.14 &0.665 \\
         ~ & &w/ ours &\textbf{26.40} &\textbf{0.681} \\ \hline
    \end{tabular}
    \vskip -0.10cm
\end{table}
\vskip -0.3cm
\begin{table}[h]
    \centering
    \caption{\textbf{Generative IR with our methods}}
    \vskip -0.3cm
    \scalebox{0.8}{
    \begin{tabular}{c|c|c|c|c}
        \hline
         model &PSNR$\uparrow$ &SSIM$\uparrow$&LPIPS$\downarrow$ &MANIQA$\uparrow$\\ \hline
         ResShift~\cite{yue2023resshift} &22.55 &0.6732&0.40 &0.2906 \\
         ResShift+ours &\textbf{22.69} &\textbf{0.6754}&\textbf{0.38} &\textbf{0.3211}\\ \hline
         RealEsrgan~\cite{wang2021real} &23.19 &0.6812 &0.31 &0.2549\\
         RealEsrgan+ours &\textbf{23.64} &\textbf{0.6881}&\textbf{0.24} &\textbf{0.2560}\\
    \end{tabular}}
    \label{tab:gen_IR}
    \vspace{-0.35cm}
\end{table}
\begin{table}[h]
    \centering
    \small
  \setlength{\abovecaptionskip}{-0pt}
	\setlength{\belowcaptionskip}{-4pt}
	\renewcommand{\arraystretch}{0.9}
     \caption{\textbf{Ablation Study.} The left table demonstrates the effectiveness of our method's components. The right table indicates the optimal hyperparameters for Implicit Adaptive Dropout.}
          \label{tab:ablation}
 \resizebox{0.9\linewidth}{!}{
 \begin{tabular}{cc}
    \begin{tabular}{@{\hspace{2pt}}ccc@{\hspace{2pt}}}
        \hline
        dropout & strategy &PSNR\\
        \hline
        None &\ding{55}  &25.66 \\
        standard &\ding{55}  &24.33 \\
        standard &\ding{51} &24.89 \\
        adaptive &\ding{55}  &25.89 \\
        adaptive &\ding{51} &\textbf{26.07} \\
        \hline
    \end{tabular}
    &
    \begin{tabular}{@{\hspace{2pt}}ccc@{\hspace{2pt}}}
        \hline
        $w$ &SRResNet &RRDB\\
        \hline
        0.5 &25.83 &26.09 \\
        0.6 &25.94 &26.22 \\
        0.7 &\textbf{26.07} &26.43 \\
        0.8 &26.00 &\textbf{26.59} \\
        0.9 &25.86 &26.50 \\
        \hline
    \end{tabular}
\end{tabular}}
\vspace{-2mm}
\end{table}
\begin{figure}[t]
\centering
\includegraphics[width=1\linewidth]{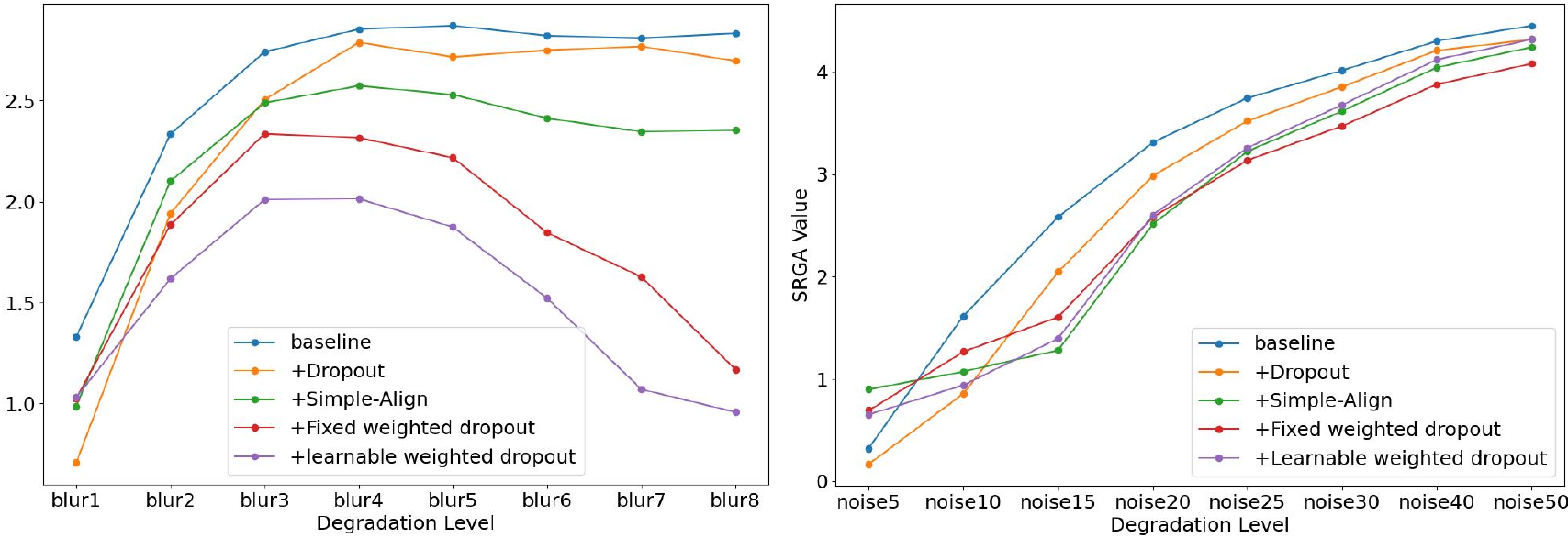}
\vspace{-5mm}
\caption{\textbf{Quantitative generalization performance compared with other regularization methods. }}
\vspace{-5mm}
\label{fig:SRGA}
\end{figure}

\noindent\textbf{Ablation study.} We conduct ablation study on different components of Adaptive Dropout as shown in~\tablename~\ref{tab:ablation}. It can be observed that both the adaptive dropout format and the adaptive training strategy contribute to improving the model's generalization capability. Moreover, we select the optimal $w$ for Implicit Adaptive Dropout. More results of ablation study will be presented in supplementary materials.

\noindent\textbf{Quantitative analysis of generalization performance.} 
To further compare generalization performance, we use SRGA~\cite{liu2023evaluating}, the first generalization performance metric in the low-level domain, to demonstrate the effectiveness of our proposed method. We show the SRGA under different levels of degradation in \figurename~\ref{fig:SRGA}, where a lower SRGA indicates better generalization performance. Our proposed method achieves lower SRGA on most degradations.

\section{Conclusion}
\label{sec:conclusion}
In this work, we first focus on the need for explicit regularization of intermediate layers in BSR networks to enhance generalization performance, and specifically analyze why dropout cannot be directly applied to intermediate layers. Subsequently, we propose Adaptive Dropout, a regularization method that can be applied anywhere in the network, further improving the network's generalization performance. We hope that this research will draw more attention to the training strategies for SR tasks.

\section{Acknowledgement}
This work was supported by the Anhui Provincial Natural Science Foundation under Grant 2108085UD12. We acknowledge the support of GPU cluster built by MCC Lab of Information Science and Technology Institution, USTC.

{
    \small
    \bibliographystyle{ieeenat_fullname}
    \bibliography{main}

\begin{thebibliography}{70}
\providecommand{\natexlab}[1]{#1}
\providecommand{\url}[1]{\texttt{#1}}
\expandafter\ifx\csname urlstyle\endcsname\relax
  \providecommand{\doi}[1]{doi: #1}\else
  \providecommand{\doi}{doi: \begingroup \urlstyle{rm}\Url}\fi

\bibitem[Agustsson and Timofte(2017)]{Agustsson_2017_CVPR_Workshops}
Eirikur Agustsson and Radu Timofte.
\newblock Ntire 2017 challenge on single image super-resolution: Dataset and
  study.
\newblock In \emph{The IEEE Conference on Computer Vision and Pattern
  Recognition (CVPR) Workshops}, 2017.

\bibitem[Bell-Kligler et~al.(2019)Bell-Kligler, Shocher, and
  Irani]{bell2019blind}
Sefi Bell-Kligler, Assaf Shocher, and Michal Irani.
\newblock Blind super-resolution kernel estimation using an internal-gan.
\newblock \emph{Advances in Neural Information Processing Systems}, 32, 2019.

\bibitem[Bello et~al.(2021)Bello, Fedus, Du, Cubuk, Srinivas, Lin, Shlens, and
  Zoph]{bello2021revisiting}
Irwan Bello, William Fedus, Xianzhi Du, Ekin~Dogus Cubuk, Aravind Srinivas,
  Tsung-Yi Lin, Jonathon Shlens, and Barret Zoph.
\newblock Revisiting resnets: Improved training and scaling strategies.
\newblock \emph{Advances in Neural Information Processing Systems},
  34:\penalty0 22614--22627, 2021.

\bibitem[Bevilacqua et~al.(2012)Bevilacqua, Roumy, Guillemot, and
  Alberi-Morel]{bevilacqua2012low}
Marco Bevilacqua, Aline Roumy, Christine Guillemot, and Marie~Line
  Alberi-Morel.
\newblock Low-complexity single-image super-resolution based on nonnegative
  neighbor embedding.
\newblock \emph{British Machine Vision Conference}, 2012.

\bibitem[Bjorck et~al.(2018)Bjorck, Gomes, Selman, and
  Weinberger]{bjorck2018understanding}
Nils Bjorck, Carla~P Gomes, Bart Selman, and Kilian~Q Weinberger.
\newblock Understanding batch normalization.
\newblock \emph{Advances in Neural Information Processing Systems}, 31, 2018.

\bibitem[Cai et~al.(2019)Cai, Zeng, Yong, Cao, and Zhang]{cai2019toward}
Jianrui Cai, Hui Zeng, Hongwei Yong, Zisheng Cao, and Lei Zhang.
\newblock Toward real-world single image super-resolution: A new benchmark and
  a new model.
\newblock In \emph{Proceedings of the IEEE/CVF International Conference on
  Computer Vision}, pages 3086--3095, 2019.

\bibitem[Chen et~al.(2022)Chen, Shi, Qin, Li, Han, Yang, and Guo]{chen2022real}
Chaofeng Chen, Xinyu Shi, Yipeng Qin, Xiaoming Li, Xiaoguang Han, Tao Yang, and
  Shihui Guo.
\newblock Real-world blind super-resolution via feature matching with implicit
  high-resolution priors.
\newblock In \emph{Proceedings of the 30th ACM International Conference on
  Multimedia}, pages 1329--1338, 2022.

\bibitem[Chen et~al.(2023{\natexlab{a}})Chen, Gu, Liu, Magid, Dong, Wang,
  Pfister, and Zhu]{chen2023masked}
Haoyu Chen, Jinjin Gu, Yihao Liu, Salma~Abdel Magid, Chao Dong, Qiong Wang,
  Hanspeter Pfister, and Lei Zhu.
\newblock Masked image training for generalizable deep image denoising.
\newblock In \emph{Proceedings of the IEEE/CVF Conference on Computer Vision
  and Pattern Recognition}, pages 1692--1703, 2023{\natexlab{a}}.

\bibitem[Chen et~al.(2023{\natexlab{b}})Chen, Wang, Zhou, Qiao, and
  Dong]{chen2023activating}
Xiangyu Chen, Xintao Wang, Jiantao Zhou, Yu Qiao, and Chao Dong.
\newblock Activating more pixels in image super-resolution transformer.
\newblock In \emph{Proceedings of the IEEE/CVF Conference on Computer Vision
  and Pattern Recognition}, pages 22367--22377, 2023{\natexlab{b}}.

\bibitem[Diederik(2014)]{diederik2014adam}
P~Kingma Diederik.
\newblock Adam: A method for stochastic optimization.
\newblock \emph{(No Title)}, 2014.

\bibitem[Dong et~al.(2022)Dong, Mou, Zhang, Fu, and Zhu]{dong2022real}
Runmin Dong, Lichao Mou, Lixian Zhang, Haohuan Fu, and Xiao~Xiang Zhu.
\newblock Real-world remote sensing image super-resolution via a practical
  degradation model and a kernel-aware network.
\newblock \emph{ISPRS Journal of Photogrammetry and Remote Sensing},
  191:\penalty0 155--170, 2022.

\bibitem[Ghiasi et~al.(2018)Ghiasi, Lin, and Le]{ghiasi2018dropblock}
Golnaz Ghiasi, Tsung-Yi Lin, and Quoc~V Le.
\newblock Dropblock: A regularization method for convolutional networks.
\newblock \emph{Advances in Neural Information Processing Systems}, 31, 2018.

\bibitem[Gu et~al.(2019)Gu, Lu, Zuo, and Dong]{gu2019blind}
Jinjin Gu, Hannan Lu, Wangmeng Zuo, and Chao Dong.
\newblock Blind super-resolution with iterative kernel correction.
\newblock In \emph{Proceedings of the IEEE/CVF Conference on Computer Vision
  and Pattern Recognition}, pages 1604--1613, 2019.

\bibitem[He et~al.(2022)He, Chen, Xie, Li, Doll{\'a}r, and
  Girshick]{he2022masked}
Kaiming He, Xinlei Chen, Saining Xie, Yanghao Li, Piotr Doll{\'a}r, and Ross
  Girshick.
\newblock Masked autoencoders are scalable vision learners.
\newblock In \emph{Proceedings of the IEEE/CVF Conference on Computer Vision
  and Pattern Recognition}, pages 16000--16009, 2022.

\bibitem[Hsu et~al.(2024)Hsu, Lee, and Chou]{hsu2024drct}
Chih-Chung Hsu, Chia-Ming Lee, and Yi-Shiuan Chou.
\newblock Drct: Saving image super-resolution away from information bottleneck.
\newblock \emph{arXiv preprint arXiv:2404.00722}, 2024.

\bibitem[Hu et~al.(2018)Hu, Shen, and Sun]{hu2018squeeze}
Jie Hu, Li Shen, and Gang Sun.
\newblock Squeeze-and-excitation networks.
\newblock In \emph{Proceedings of the IEEE/CVF Conference on Computer Vision
  and Pattern Recognition}, pages 7132--7141, 2018.

\bibitem[Huang et~al.(2015)Huang, Singh, and Ahuja]{huang2015single}
Jia-Bin Huang, Abhishek Singh, and Narendra Ahuja.
\newblock Single image super-resolution from transformed self-exemplars.
\newblock In \emph{Proceedings of the IEEE/CVF Conference on Computer Vision
  and Pattern Recognition}, pages 5197--5206, 2015.

\bibitem[Huang et~al.(2020)Huang, Li, Wang, Tan, et~al.]{huang2020unfolding}
Yan Huang, Shang Li, Liang Wang, Tieniu Tan, et~al.
\newblock Unfolding the alternating optimization for blind super resolution.
\newblock \emph{Advances in Neural Information Processing Systems},
  33:\penalty0 5632--5643, 2020.

\bibitem[Huang et~al.(2022)Huang, Fu, Li, and Zha]{huang2022learning}
Yukun Huang, Xueyang Fu, Liang Li, and Zheng-Jun Zha.
\newblock Learning degradation-invariant representation for robust real-world
  person re-identification.
\newblock \emph{International Journal of Computer Vision}, 130\penalty0
  (11):\penalty0 2770--2796, 2022.

\bibitem[Ji et~al.(2022)Ji, Wang, Tao, Huang, Hua, and Lu]{sstkd}
Deyi Ji, Haoran Wang, Mingyuan Tao, Jianqiang Huang, Xian-Sheng Hua, and
  Hongtao Lu.
\newblock Structural and statistical texture knowledge distillation for
  semantic segmentation.
\newblock In \emph{Proceedings of the IEEE/CVF Conference on Computer Vision
  and Pattern Recognition}, pages 16876--16885, 2022.

\bibitem[Ji et~al.(2023{\natexlab{a}})Ji, Zhao, and Lu]{gpwformer}
Deyi Ji, Feng Zhao, and Hongtao Lu.
\newblock Guided patch-grouping wavelet transformer with spatial congruence for
  ultra-high resolution segmentation.
\newblock \emph{International Joint Conference on Artificial Intelligence},
  pages 920--928, 2023{\natexlab{a}}.

\bibitem[Ji et~al.(2023{\natexlab{b}})Ji, Zhao, Lu, Tao, and Ye]{urur}
Deyi Ji, Feng Zhao, Hongtao Lu, Mingyuan Tao, and Jieping Ye.
\newblock Ultra-high resolution segmentation with ultra-rich context: A novel
  benchmark.
\newblock In \emph{Proceedings of the IEEE/CVF Conference on Computer Vision
  and Pattern Recognition}, pages 23621--23630, 2023{\natexlab{b}}.

\bibitem[Ji et~al.(2024{\natexlab{a}})Ji, Jin, Lu, and Zhao]{pptformer}
Deyi Ji, Wenwei Jin, Hongtao Lu, and Feng Zhao.
\newblock Pptformer: Pseudo multi-perspective transformer for uav segmentation.
\newblock \emph{International Joint Conference on Artificial Intelligence},
  pages 893--901, 2024{\natexlab{a}}.

\bibitem[Ji et~al.(2024{\natexlab{b}})Ji, Zhao, Zhu, Jin, Lu, and Ye]{dlpl}
Deyi Ji, Feng Zhao, Lanyun Zhu, Wenwei Jin, Hongtao Lu, and Jieping Ye.
\newblock Discrete latent perspective learning for segmentation and detection.
\newblock In \emph{International Conference on Machine Learning}, pages
  21719--21730, 2024{\natexlab{b}}.

\bibitem[Ji et~al.(2025)Ji, Zhao, Lu, Wu, and Ye]{sstkd_pami}
Deyi Ji, Feng Zhao, Hongtao Lu, Feng Wu, and Jieping Ye.
\newblock Structural and statistical texture knowledge distillation and
  learning for segmentation.
\newblock \emph{IEEE Transactions on Pattern Analysis and Machine
  Intelligence}, pages 1--18, 2025.

\bibitem[Jiang et~al.(2020)Jiang, Wang, Yi, Chen, Huang, Luo, Ma, and
  Jiang]{jiang2020multi}
Kui Jiang, Zhongyuan Wang, Peng Yi, Chen Chen, Baojin Huang, Yimin Luo, Jiayi
  Ma, and Junjun Jiang.
\newblock Multi-scale progressive fusion network for single image deraining.
\newblock In \emph{Proceedings of the IEEE/CVF Conference on Computer Vision
  and Pattern Recognition}, pages 8346--8355, 2020.

\bibitem[Johnson and Zhang(2024)]{johnson2024inconsistency}
Rie Johnson and Tong Zhang.
\newblock Inconsistency, instability, and generalization gap of deep neural
  network training.
\newblock \emph{Advances in Neural Information Processing Systems}, 36, 2024.

\bibitem[Khan et~al.(2024)Khan, Micheloni, and Martinel]{khan2024lightweight}
Asif~Hussain Khan, Christian Micheloni, and Niki Martinel.
\newblock Lightweight prompt learning implicit degradation estimation network
  for blind super resolution.
\newblock \emph{IEEE Transactions on Image Processing}, 2024.

\bibitem[Kong et~al.(2022)Kong, Liu, Gu, Qiao, and Dong]{kong2022reflash}
Xiangtao Kong, Xina Liu, Jinjin Gu, Yu Qiao, and Chao Dong.
\newblock Reflash dropout in image super-resolution.
\newblock In \emph{Proceedings of the IEEE/CVF Conference on Computer Vision
  and Pattern Recognition}, pages 6002--6012, 2022.

\bibitem[Ledig et~al.(2017)Ledig, Theis, Husz{\'a}r, Caballero, Cunningham,
  Acosta, Aitken, Tejani, Totz, Wang, et~al.]{ledig2017photo}
Christian Ledig, Lucas Theis, Ferenc Husz{\'a}r, Jose Caballero, Andrew
  Cunningham, Alejandro Acosta, Andrew Aitken, Alykhan Tejani, Johannes Totz,
  Zehan Wang, et~al.
\newblock Photo-realistic single image super-resolution using a generative
  adversarial network.
\newblock In \emph{Proceedings of the IEEE/CVF Conference on Computer Vision
  and Pattern Recognition}, pages 4681--4690, 2017.

\bibitem[Li et~al.(2019{\natexlab{a}})Li, Ren, Fu, Tao, Feng, Zeng, and
  Wang]{li2019benchmarking}
Boyi Li, Wenqi Ren, Dengpan Fu, Dacheng Tao, Dan Feng, Wenjun Zeng, and
  Zhangyang Wang.
\newblock Benchmarking single-image dehazing and beyond.
\newblock \emph{IEEE Transactions on Image Processing}, 28\penalty0
  (1):\penalty0 492--505, 2019{\natexlab{a}}.

\bibitem[Li et~al.(2022)Li, Wu, Bai, Lin, Cong, and Zhao]{li2022learning}
Feng Li, Yixuan Wu, Huihui Bai, Weisi Lin, Runmin Cong, and Yao Zhao.
\newblock Learning detail-structure alternative optimization for blind
  super-resolution.
\newblock \emph{IEEE Transactions on Multimedia}, 25:\penalty0 2825--2838,
  2022.

\bibitem[Li et~al.(2019{\natexlab{b}})Li, Chen, Hu, and
  Yang]{li2019understanding}
Xiang Li, Shuo Chen, Xiaolin Hu, and Jian Yang.
\newblock Understanding the disharmony between dropout and batch normalization
  by variance shift.
\newblock In \emph{Proceedings of the IEEE/CVF Conference on Computer Vision
  and Pattern Recognition}, pages 2682--2690, 2019{\natexlab{b}}.

\bibitem[Liang et~al.(2021{\natexlab{a}})Liang, Cao, Sun, Zhang, Van~Gool, and
  Timofte]{liang2021swinir}
Jingyun Liang, Jiezhang Cao, Guolei Sun, Kai Zhang, Luc Van~Gool, and Radu
  Timofte.
\newblock Swinir: Image restoration using swin transformer.
\newblock In \emph{Proceedings of the IEEE/CVF International Conference on
  Computer Vision}, pages 1833--1844, 2021{\natexlab{a}}.

\bibitem[Liang et~al.(2021{\natexlab{b}})Liang, Zhang, Gu, Van~Gool, and
  Timofte]{liang2021flow}
Jingyun Liang, Kai Zhang, Shuhang Gu, Luc Van~Gool, and Radu Timofte.
\newblock Flow-based kernel prior with application to blind super-resolution.
\newblock In \emph{Proceedings of the IEEE/CVF Conference on Computer Vision
  and Pattern Recognition}, pages 10601--10610, 2021{\natexlab{b}}.

\bibitem[Liang et~al.(2022)Liang, Zeng, and Zhang]{liang2022efficient}
Jie Liang, Hui Zeng, and Lei Zhang.
\newblock Efficient and degradation-adaptive network for real-world image
  super-resolution.
\newblock In \emph{European Conference on Computer Vision}, pages 574--591.
  Springer, 2022.

\bibitem[Liu et~al.(2022{\natexlab{a}})Liu, Liu, Gu, Qiao, and
  Dong]{liu2022blind}
Anran Liu, Yihao Liu, Jinjin Gu, Yu Qiao, and Chao Dong.
\newblock Blind image super-resolution: A survey and beyond.
\newblock \emph{IEEE Transactions on Pattern Analysis and Machine
  Intelligence}, 45\penalty0 (5):\penalty0 5461--5480, 2022{\natexlab{a}}.

\bibitem[Liu et~al.(2022{\natexlab{b}})Liu, Li, Fu, Xiao, Gao, Hu, and
  Liu]{liu2022degradation}
Haofeng Liu, Heng Li, Huazhu Fu, Ruoxiu Xiao, Yunshu Gao, Yan Hu, and Jiang
  Liu.
\newblock Degradation-invariant enhancement of fundus images via pyramid
  constraint network.
\newblock In \emph{International Conference on Medical Image Computing and
  Computer-Assisted Intervention}, pages 507--516. Springer,
  2022{\natexlab{b}}.

\bibitem[Liu et~al.(2023)Liu, Zhao, Gu, Qiao, and Dong]{liu2023evaluating}
Yihao Liu, Hengyuan Zhao, Jinjin Gu, Yu Qiao, and Chao Dong.
\newblock Evaluating the generalization ability of super-resolution networks.
\newblock \emph{IEEE Transactions on pattern analysis and machine
  intelligence}, 2023.

\bibitem[Long et~al.(2018)Long, Cao, Cao, Wang, and
  Jordan]{long2018transferable}
Mingsheng Long, Yue Cao, Zhangjie Cao, Jianmin Wang, and Michael~I Jordan.
\newblock Transferable representation learning with deep adaptation networks.
\newblock \emph{IEEE Transactions on Pattern Analysis and Machine
  Intelligence}, 41\penalty0 (12):\penalty0 3071--3085, 2018.

\bibitem[Martin et~al.(2001)Martin, Fowlkes, Tal, and
  Malik]{martin2001database}
David Martin, Charless Fowlkes, Doron Tal, and Jitendra Malik.
\newblock A database of human segmented natural images and its application to
  evaluating segmentation algorithms and measuring ecological statistics.
\newblock In \emph{Proceedings eighth IEEE international conference on computer
  vision. ICCV 2001}, pages 416--423. IEEE, 2001.

\bibitem[Matsui et~al.(2017)Matsui, Ito, Aramaki, Fujimoto, Ogawa, Yamasaki,
  and Aizawa]{matsui2017sketch}
Yusuke Matsui, Kota Ito, Yuji Aramaki, Azuma Fujimoto, Toru Ogawa, Toshihiko
  Yamasaki, and Kiyoharu Aizawa.
\newblock Sketch-based manga retrieval using manga109 dataset.
\newblock \emph{Multimedia tools and applications}, 76:\penalty0 21811--21838,
  2017.

\bibitem[Michaeli and Irani(2013)]{michaeli2013nonparametric}
Tomer Michaeli and Michal Irani.
\newblock Nonparametric blind super-resolution.
\newblock In \emph{Proceedings of the IEEE International Conference on Computer
  Vision}, pages 945--952, 2013.

\bibitem[Morcos et~al.(2018)Morcos, Barrett, Rabinowitz, and
  Botvinick]{morcos2018importance}
Ari~S Morcos, David~GT Barrett, Neil~C Rabinowitz, and Matthew Botvinick.
\newblock On the importance of single directions for generalization.
\newblock \emph{arXiv preprint arXiv:1803.06959}, 2018.

\bibitem[Neyshabur(2017)]{neyshabur2017implicit}
Behnam Neyshabur.
\newblock Implicit regularization in deep learning.
\newblock \emph{arXiv preprint arXiv:1709.01953}, 2017.

\bibitem[Neyshabur et~al.(2014)Neyshabur, Tomioka, and
  Srebro]{neyshabur2014search}
Behnam Neyshabur, Ryota Tomioka, and Nathan Srebro.
\newblock In search of the real inductive bias: On the role of implicit
  regularization in deep learning.
\newblock \emph{arXiv preprint arXiv:1412.6614}, 2014.

\bibitem[Oh et~al.(2023)Oh, Kim, Lee, and Chung]{oh2023super}
Han Oh, Dongjin Kim, Sun~Gu Lee, and Daewon Chung.
\newblock Super-resolution of remote sensing imagery using implicit degradation
  modeling.
\newblock In \emph{IGARSS 2023-2023 IEEE International Geoscience and Remote
  Sensing Symposium}, pages 5146--5149. IEEE, 2023.

\bibitem[{\"O}zg{\"u}r and Nar(2020)]{ozgur2020effect}
Atilla {\"O}zg{\"u}r and Fatih Nar.
\newblock Effect of dropout layer on classical regression problems.
\newblock In \emph{2020 28th Signal Processing and Communications Applications
  Conference (SIU)}, pages 1--4. IEEE, 2020.

\bibitem[Qian et~al.(2022)Qian, Li, Peng, Mai, Hammoud, Elhoseiny, and
  Ghanem]{qian2022pointnext}
Guocheng Qian, Yuchen Li, Houwen Peng, Jinjie Mai, Hasan Hammoud, Mohamed
  Elhoseiny, and Bernard Ghanem.
\newblock Pointnext: Revisiting pointnet++ with improved training and scaling
  strategies.
\newblock \emph{Advances in Neural Information Processing Systems},
  35:\penalty0 23192--23204, 2022.

\bibitem[Qin et~al.(2020)Qin, Wang, Bai, Xie, and Jia]{qin2020ffa}
Xu Qin, Zhilin Wang, Yuanchao Bai, Xiaodong Xie, and Huizhu Jia.
\newblock Ffa-net: Feature fusion attention network for single image dehazing.
\newblock In \emph{Proceedings of the AAAI Conference on Artificial
  Intelligence}, pages 11908--11915, 2020.

\bibitem[Qin et~al.(2024)Qin, Nie, Wang, Liu, Sun, Zhu, Lu, and
  Pan]{qin2024multi}
Yi Qin, Haitao Nie, Jiarong Wang, Huiying Liu, Jiaqi Sun, Ming Zhu, Jie Lu, and
  Qi Pan.
\newblock Multi-degradation super-resolution reconstruction for remote sensing
  images with reconstruction features-guided kernel correction.
\newblock \emph{Remote Sensing}, 16\penalty0 (16):\penalty0 2915, 2024.

\bibitem[Sahak et~al.(2023)Sahak, Watson, Saharia, and
  Fleet]{sahak2023denoising}
Hshmat Sahak, Daniel Watson, Chitwan Saharia, and David Fleet.
\newblock Denoising diffusion probabilistic models for robust image
  super-resolution in the wild.
\newblock \emph{arXiv preprint arXiv:2302.07864}, 2023.

\bibitem[Srivastava et~al.(2014)Srivastava, Hinton, Krizhevsky, Sutskever, and
  Salakhutdinov]{srivastava2014dropout}
Nitish Srivastava, Geoffrey Hinton, Alex Krizhevsky, Ilya Sutskever, and Ruslan
  Salakhutdinov.
\newblock Dropout: a simple way to prevent neural networks from overfitting.
\newblock \emph{The journal of machine learning research}, 15\penalty0
  (1):\penalty0 1929--1958, 2014.

\bibitem[Timofte et~al.(2017)Timofte, Agustsson, Van~Gool, Yang, and
  Zhang]{timofte2017ntire}
Radu Timofte, Eirikur Agustsson, Luc Van~Gool, Ming-Hsuan Yang, and Lei Zhang.
\newblock Ntire 2017 challenge on single image super-resolution: Methods and
  results.
\newblock In \emph{Proceedings of the IEEE/CVF Conference on Computer Vision
  and Pattern Recognition workshops}, pages 114--125, 2017.

\bibitem[Wang et~al.(2023)Wang, Xie, Zhao, Li, Liang, Zheng, and
  Meng]{wang2023rcdnet}
Hong Wang, Qi Xie, Qian Zhao, Yuexiang Li, Yong Liang, Yefeng Zheng, and Deyu
  Meng.
\newblock Rcdnet: An interpretable rain convolutional dictionary network for
  single image deraining.
\newblock \emph{IEEE Transactions on Neural Networks and Learning Systems},
  2023.

\bibitem[Wang et~al.(2024)Wang, Chen, Zheng, and Zeng]{wang2024navigating}
Hongjun Wang, Jiyuan Chen, Yinqiang Zheng, and Tieyong Zeng.
\newblock Navigating beyond dropout: An intriguing solution towards
  generalizable image super resolution.
\newblock In \emph{Proceedings of the IEEE/CVF Conference on Computer Vision
  and Pattern Recognition}, pages 25532--25543, 2024.

\bibitem[Wang et~al.(2021)Wang, Xie, Dong, and Shan]{wang2021real}
Xintao Wang, Liangbin Xie, Chao Dong, and Ying Shan.
\newblock Real-esrgan: Training real-world blind super-resolution with pure
  synthetic data.
\newblock In \emph{Proceedings of the IEEE/CVF International Conference on
  Computer Vision}, pages 1905--1914, 2021.

\bibitem[Wei et~al.(2020)Wei, Xie, Lu, Zhan, Ye, Zuo, and
  Lin]{wei2020component}
Pengxu Wei, Ziwei Xie, Hannan Lu, Zongyuan Zhan, Qixiang Ye, Wangmeng Zuo, and
  Liang Lin.
\newblock Component divide-and-conquer for real-world image super-resolution.
\newblock In \emph{Computer Vision--ECCV 2020: 16th European Conference,
  Glasgow, UK, August 23--28, 2020, Proceedings, Part VIII 16}, pages 101--117.
  Springer, 2020.

\bibitem[Wu et~al.(2021)Wu, Li, Wang, Meng, Qin, Chen, Zhang, Liu,
  et~al.]{wu2021r}
Lijun Wu, Juntao Li, Yue Wang, Qi Meng, Tao Qin, Wei Chen, Min Zhang, Tie-Yan
  Liu, et~al.
\newblock R-drop: Regularized dropout for neural networks.
\newblock \emph{Advances in Neural Information Processing Systems},
  34:\penalty0 10890--10905, 2021.

\bibitem[Wu and He(2018)]{wu2018group}
Yuxin Wu and Kaiming He.
\newblock Group normalization.
\newblock In \emph{Proceedings of the European Conference on Computer Vision
  (ECCV)}, pages 3--19, 2018.

\bibitem[Xu et~al.(2019)Xu, Sun, Zhang, Zhao, and Lin]{xu2019understanding}
Jingjing Xu, Xu Sun, Zhiyuan Zhang, Guangxiang Zhao, and Junyang Lin.
\newblock Understanding and improving layer normalization.
\newblock \emph{Advances in Neural Information Processing Systems}, 32, 2019.

\bibitem[Yan et~al.(2024)Yan, Niu, Wang, Dong, Wo{\'z}niak, and
  Zhang]{yan2024kgsr}
Qingsen Yan, Axi Niu, Chaoqun Wang, Wei Dong, Marcin Wo{\'z}niak, and Yanning
  Zhang.
\newblock Kgsr: A kernel guided network for real-world blind super-resolution.
\newblock \emph{Pattern Recognition}, 147:\penalty0 110095, 2024.

\bibitem[Yang et~al.(2010)Yang, Wright, Huang, and Ma]{yang2010image}
Jianchao Yang, John Wright, Thomas~S Huang, and Yi Ma.
\newblock Image super-resolution via sparse representation.
\newblock \emph{IEEE Transactions on Image Processing}, 19\penalty0
  (11):\penalty0 2861--2873, 2010.

\bibitem[Yosinski et~al.(2014)Yosinski, Clune, Bengio, and
  Lipson]{yosinski2014transferable}
Jason Yosinski, Jeff Clune, Yoshua Bengio, and Hod Lipson.
\newblock How transferable are features in deep neural networks?
\newblock \emph{Advances in Neural Information Processing Systems}, 27, 2014.

\bibitem[Yue et~al.(2022)Yue, Zhao, Xie, Zhang, Meng, and Wong]{yue2022blind}
Zongsheng Yue, Qian Zhao, Jianwen Xie, Lei Zhang, Deyu Meng, and Kwan-Yee~K
  Wong.
\newblock Blind image super-resolution with elaborate degradation modeling on
  noise and kernel.
\newblock In \emph{Proceedings of the IEEE/CVF Conference on Computer Vision
  and Pattern Recognition}, pages 2128--2138, 2022.

\bibitem[Yue et~al.(2023)Yue, Wang, and Loy]{yue2023resshift}
Zongsheng Yue, Jianyi Wang, and Chen~Change Loy.
\newblock Resshift: Efficient diffusion model for image super-resolution by
  residual shifting.
\newblock \emph{Advances in Neural Information Processing Systems},
  36:\penalty0 13294--13307, 2023.

\bibitem[Zamir et~al.(2022)Zamir, Arora, Khan, Hayat, Khan, and
  Yang]{zamir2022restormer}
Syed~Waqas Zamir, Aditya Arora, Salman Khan, Munawar Hayat, Fahad~Shahbaz Khan,
  and Ming-Hsuan Yang.
\newblock Restormer: Efficient transformer for high-resolution image
  restoration.
\newblock In \emph{Proceedings of the IEEE/CVF Conference on Computer Vision
  and Pattern Recognition}, pages 5728--5739, 2022.

\bibitem[Zhang et~al.(2021)Zhang, Liang, Van~Gool, and
  Timofte]{zhang2021designing}
Kai Zhang, Jingyun Liang, Luc Van~Gool, and Radu Timofte.
\newblock Designing a practical degradation model for deep blind image
  super-resolution.
\newblock In \emph{Proceedings of the IEEE/CVF International Conference on
  Computer Vision}, pages 4791--4800, 2021.

\bibitem[Zhang et~al.(2018)Zhang, Li, Li, Wang, Zhong, and Fu]{zhang2018image}
Yulun Zhang, Kunpeng Li, Kai Li, Lichen Wang, Bineng Zhong, and Yun Fu.
\newblock Image super-resolution using very deep residual channel attention
  networks.
\newblock In \emph{Proceedings of the European Conference on Computer Vision
  (ECCV)}, pages 286--301, 2018.

\bibitem[Zhang et~al.(2022)Zhang, Dong, Yang, Qing, He, and
  Chen]{zhang2022weakly}
Yongfei Zhang, Ling Dong, Hong Yang, Linbo Qing, Xiaohai He, and Honggang Chen.
\newblock Weakly-supervised contrastive learning-based implicit degradation
  modeling for blind image super-resolution.
\newblock \emph{Knowledge-Based Systems}, 249:\penalty0 108984, 2022.

\end{thebibliography}
}


\end{document}